%% file: paper.tex
\PassOptionsToPackage{unicode}{hyperref}
\PassOptionsToPackage{hyphens}{url}
\documentclass[
  journal]{IEEEtran}
\usepackage{xcolor}
\usepackage{amsmath,amssymb}
\usepackage{iftex}
\ifPDFTeX
  \usepackage[T1]{fontenc}
  \usepackage[utf8]{inputenc}
  \usepackage{textcomp}
\else
  \usepackage{unicode-math}
  \defaultfontfeatures{Scale=MatchLowercase}
  \defaultfontfeatures[\rmfamily]{Ligatures=TeX,Scale=1}
\fi
\usepackage{lmodern}
\ifPDFTeX\else
\fi
\IfFileExists{upquote.sty}{\usepackage{upquote}}{}
\IfFileExists{microtype.sty}{%
  \usepackage[]{microtype}
  \UseMicrotypeSet[protrusion]{basicmath}
}{}
\makeatletter
\@ifundefined{KOMAClassName}{%
  \IfFileExists{parskip.sty}{%
    \usepackage{parskip}
  }{%
    \setlength{\parindent}{0pt}
    \setlength{\parskip}{6pt plus 2pt minus 1pt}}
}{%
  \KOMAoptions{parskip=half}}
\makeatother
\providecommand{\tightlist}{%
  \setlength{\itemsep}{0pt}\setlength{\parskip}{0pt}}
\usepackage[numbers]{natbib}
\usepackage[most]{tcolorbox}
\usepackage{booktabs}
\usepackage{array}
\usepackage{graphicx}

\title{When GeoFMs Meet VHR: Benchmarking Foundation Models on Very High Resolution Earth Observation Imagery}
\author{Kshitij Raj Sharma\textsuperscript{*1}, Pierre Adorni\textsuperscript{*1}, Minh-Tan Pham\textsuperscript{1}, Dirk Tiede\textsuperscript{2}, S\'{e}bastien Lef\`{e}vre\textsuperscript{1,3}\thanks{\textsuperscript{*}Equal contribution. \textsuperscript{1}Universit\'{e} Bretagne Sud, IRISA, France. \textsuperscript{2}University of Salzburg, Department of Geoinformatics, Austria. \textsuperscript{3}UiT The Arctic University of Norway, Troms\o, Norway. Corresponding author: skshitizraj@gmail.com}}
\usepackage{bookmark}
\IfFileExists{xurl.sty}{\usepackage{xurl}}{}
\hypersetup{
  hidelinks}
\usepackage[capitalise]{cleveref}

\begin{document}

\maketitle

\bstctlcite{BSTctl}

\begin{abstract}
Geospatial foundation models are a widely adopted starting point for remote sensing tasks. Most are pretrained on Sentinel-1/2 and Landsat imagery at a ground sampling distance (GSD) of 10m or coarser. Many applications instead rely on very high resolution (VHR) imagery, where individual buildings, roads, and amenities are resolved. We study which pretraining regimes transfer well to this scale. We introduce VHR-Bench, a benchmark suite of ten datasets at 1.5m GSD or finer spanning five task families, eight application domains, and both satellite and aerial platforms. We evaluate twelve encoders in three pretraining tiers with frozen backbones at two training-data fractions. We find that both pretraining on large natural-image collections and VHR-specific pretraining transfer effectively to VHR tasks, with complementary strengths. SkySense leads when labels are scarce; at the full label fraction the web-pretrained DINOv3 ViT-L edges ahead, with the two close together there. Encoders pretrained only on Sentinel-scale imagery transfer least well under RGB input. Guided by these complementary results, we continue the self-supervised pretraining of DINOv3 ViT-L on 1.02 million held-out VHR tiles and evaluate the resulting encoder without changing the downstream protocol. The continued model gives a clear improvement over the released backbone when labels are scarce and a modest one at the full fraction, at no increase in inference parameters or compute. These findings support a generalize-then-specialize strategy: already strong, broad natural-image pretraining can be further improved by a short VHR continuation stage while preserving its general visual features.
\end{abstract}

\begin{IEEEkeywords}
Very high resolution, geospatial foundation models, benchmarking.
\end{IEEEkeywords}

\input{sec/1_intro}

\input{sec/2_related_works}
\input{sec/3_vhr}

\input{sec/4_datasets}

\input{sec/5_encoders}

\input{sec/6_protocol}
\input{sec/7_continued_pretraining}

\input{sec/8_results}

\input{sec/9_discussion}

\input{sec/11_conclusion}

\section{Acknowledgments}\label{acknowledgments}

This work has been developed in the framework of the Erasmus Mundus
Joint Master (EMJM) ``Copernicus Master in Digital Earth'', jointly
coordinated by University of Salzburg, Department of Geoinformatics,
Austria together with University of South Brittany, Mathematics, Computer
Science, Statistics Department, France and Palack\'{y} University Olomouc,
Department of Geoinformatics, Czechia.

The project was funded in part by the Université Bretagne Sud, the Centre National d'Etudes Spatiales (CNES), and the Agence Nationale de la Recherche (ANR) through two projects: DREAMS (ANR-25-CE56-4384-01) and Cluster SequoIA (ANR-23-IACL-0009). It was granted HPC computing and storage resources by GENCI-IDRIS under grant 2025-AD011015819R1 on the V100 and H100 partitions of the Jean Zay supercomputer.

The authors thank the providers of the datasets and imagery used in this
study, including SpaceNet, IGN France (FLAIR 2, PureForest, and
Open-Canopy), OpenAerialMap and the Humanitarian OpenStreetMap Team, and
the other dataset publishers. The authors also acknowledge the earlier
benchmarking efforts this work builds on, in particular the GEO-Bench,
GEO-Bench-2, and PANGAEA benchmarks. 

Generative AI was used to assist with language editing and with drafting
parts of this manuscript. All AI-assisted sections were reviewed and
verified by the authors.

\bibliography{references.bib}

\end{document}

%% file: sec/1_intro.tex
\begin{table*}[t]
\centering
\caption{Datasets in the VHR suite with sensor, platform, resolution, and region. Some datasets carry extra bands (near-infrared and elevation in FLAIR 2, near-infrared in PureForest and Open-Canopy-Small); only their RGB is evaluated.}
\label{tab:datasets}
\footnotesize
\setlength{\tabcolsep}{2pt}
\resizebox{\textwidth}{!}{%
\begin{tabular}{@{}rllllllr@{}}
\toprule
\# & Dataset & Platform & Sensor & GSD & Region & Task & Train / Val / Test \\
\midrule
1 & SpaceNet 2 \cite{van2018spacenet} & Satellite & WorldView-3 & 0.3m & 4 cities & Segmentation (buildings) & 5186 / 1461 / 2961 \\
2 & SpaceNet 3 \cite{van2018spacenet} & Satellite & WorldView-3 & 0.3m & 4 cities & Segmentation (roads) & 1782 / 510 / 257 \\
3 & xView2 (xBD) \cite{gupta2019creating} & Satellite & Mixed (Maxar) & $<$0.8m & Global & Change detection (damage) & 13384 / 917 / 933 \\
4 & FLAIR 2 \cite{ign2023flair2} & Aerial & Aerial RGBN & 0.2m & France & Segmentation (land cover) & 4049 / 1022 / 3022 \\
5 & Open-Canopy-Small \cite{fogel2024opencanopy} & Satellite & SPOT 6/7 & 1.5m & France & Regression (canopy height) & 6634 / 737 / 13675 \\
6 & EverWatch \cite{garner2024everwatch} & Aerial & Aerial RGB & 0.1m & United States & Object detection (birds) & 4429 / 500 / 196 \\
7 & NZ Cattle \cite{nz_cattle} & Aerial & Aerial RGB (LINZ) & 0.1m & New Zealand & Object detection (cattle) & 524 / 66 / 65 \\
8 & m-pv4ger \cite{mayer20223d} & Aerial & Aerial RGB & 0.1m & Germany & Classification (PV) & 11814 / 999 / 999 \\
9 & PureForest \cite{gaydon2025pureforest} & Aerial & Aerial NRGB (IGN) & 0.2m & France & Classification (tree species) & 69111 / 13523 / 52935 \\
10 & HOTOSM Buildings \cite{hot_building_segmentation_2026} & Mixed & Mixed (OpenAerialMap) & 0.3-0.5m & 15 countries & Segmentation (buildings) & 57890 / 7237 / 7236 \\
\bottomrule
\end{tabular}}
\end{table*}

\section{Introduction}\label{introduction}

Self-supervised pretraining on large satellite archives has made
Geospatial Foundation Models (GeoFMs), also referred to as Earth Observation Foundation Models (EO-FMs), a common backbone for remote
sensing tasks. Most rely on Sentinel-1 and Sentinel-2 imagery, which is openly
available on a global scale with high temporal revisit, at a Ground
Sampling Distance (GSD) of 10m or coarser. This large amount of
medium to high-resolution data has driven substantial progress in broad-area
scene analysis, but leaves an important resolution regime largely
unaddressed: Very High Resolution (VHR).

We focus on the practical demand that lies in this regime. Applications from
disaster response to infrastructure mapping depend on VHR imagery, 
acquired from satellite platforms such as
WorldView, Pléiades, and SPOT, or from aerial and drone surveys. At such resolutions, fine-grained features such as
buildings, roads, amenities, and tree crowns are discriminated at the
pixel level, enabling spatially precise tasks including building
footprint extraction, road mapping, post-disaster damage assessment,
canopy height regression, photovoltaic panel detection, tree species
classification, and many more. Many disaster tasks are time-critical and
demand object-level accurate spatial outputs.

Whether existing GeoFMs transfer well to this resolution regime has not
been systematically evaluated. Our first hypothesis is that
pretraining on medium to high-resolution Sentinel archives may be insufficient
at VHR, given the substantial scale mismatch between pretraining and
inference conditions. Conversely, we hypothesize that computer-vision
models pretrained on large RGB natural image corpora may transfer well
to optical VHR imagery: beyond sharing the same three-channel representation,
VHR imagery is closer to natural images than Sentinel-2 imagery in 
terms of spatial scale, texture, visual diversity, and fine-grained detail.
We therefore investigate whether GeoFMs without VHR pretraining underperform
on VHR tasks and whether generic RGB models provide a strong alternative to
VHR-specific pretraining.

We make three contributions. First, we introduce VHR-Bench, a benchmark of
ten VHR datasets spanning five task families and covering eight
application domains on both satellite and aerial optical sensors, under
one common evaluation framework. Second, we evaluate twelve encoders
organized into three pretraining tiers under a fixed protocol at two
training-data fractions (10 and 100\%). Third, as a design direction 
derived from the benchmark's results, and to test whether label-free adaptation
can improve an already strong natural image encoder on the VHR domain,
we assemble 1.02 million VHR tiles from seven open sources and use them to
continue DINOv3 ViT-L's self-supervised pretraining.
We will release\footnote{upon acceptance.} the resulting model and its pretraining corpus.
Together, VHR-Bench and this
evaluation give the community a shared reference for how well current
GeoFMs transfer to the VHR regime. The evaluation shows that generic
natural-image and VHR-pretrained encoders achieve the strongest performance, while Sentinel-only encoders consistently lag behind under RGB input. These findings challenge
the assumption that geospatial pretraining is inherently preferable for 
VHR imagery, and provide a basis for assessing the additional value of 
VHR-specific pretraining.

%% file: sec/2_related_works.tex
\section{Related Work}\label{related-work}

\subsection{Geospatial foundation
models}\label{geospatial-foundation-models}

A foundation model is trained on broad data, generally with
self-supervision at scale, and can be adapted to a wide range of
downstream tasks~\cite{bommasani2021, Xiao_2025}.
Geospatial foundation models apply this to Earth observation: they learn
general-purpose representations from large unlabeled Earth observation
archives typically through self-supervised objectives, then transfer 
to downstream tasks with limited supervision. 
A large part of the published models target the Sentinel and Landsat
archives where global coverage and frequent 
revisit provide plenty of pretraining data, for example Prithvi-EO
2.0~\cite{prithvi2} and TerraMind~\cite{terramind}. Others also include
higher-resolution or multi-sensor imagery, varying the GSD of their
pretraining data, such as DOFA~\cite{dofa}, Panopticon~\cite{panopticon}, 
SkySense~\cite{skysense} and EoS-FM~\cite{eosfm}. In parallel, general-purpose
computer vision models such as DINOv3~\cite{dinov3} learn transferable 
features from large web-image  collections without any remote-sensing-specific focus. 
This study benchmarks these pretraining-type families in the VHR regime.

A related line of work, represented by Tessera~\cite{tessera} and AlphaEarth Foundations~\cite{alphaearth}, distributes precomputed annual geospatial embeddings at 10m resolution rather than a model intended for direct application to imagery. Although valuable for large-scale spatial and temporal analysis, these products cannot resolve several object-level or temporally narrow VHR tasks considered here, such as counting individual cattle or detecting localized post-disaster changes, and are therefore outside the scope of this benchmark. This study instead benchmarks transferable pretraining-model families in the VHR regime.

\subsection{Benchmarks for geospatial foundation
models}\label{benchmarks-for-geospatial-foundation-models}

Evaluation of GeoFMs has moved from heterogeneous selected-dataset
studies to standardized multi-task benchmarks. The PANGAEA
benchmark~\cite{pangaea} defines a standardized
evaluation protocol spanning a diverse set of datasets, tasks,
resolutions, sensor modalities, and temporal settings, and reports that
geospatial foundation models at the time did not consistently outperform 
supervised baselines such as U-Net and a vanilla ViT.
GEO-Bench-2~\cite{geobench2} assembles nineteen
permissively licensed datasets across five task families,
classification, segmentation, regression, object detection, and instance
segmentation, and introduces capability groups that rank models on
datasets sharing characteristics such as resolution, spectral bands, and
temporality; among its findings, models pretrained on natural images
excel on high-resolution tasks, while Earth observation specific models
are stronger on multispectral applications.

Both benchmarks span a wide range of resolutions, so only a small part of
their coverage falls in the VHR regime: only four of GEO-Bench-2's nineteen
datasets are at 1.5m GSD or finer. GEO-Bench-2 partly addresses this
through its capability groups, but the VHR datasets available within these
suites are not assembled as a wide, task-balanced benchmark for VHR imagery
on their own. VHR-Bench is designed
to solve this problem and is complementary to these broader efforts.
Beyond the resolution focus, our benchmark adds a controlled comparison that
capability groups cannot provide: a matched pair of the same
architecture and training scheme (DINOv3) pretrained on web images and 
on satellite imagery, which holds the architecture fixed and shows that 
VHR-specific pretraining comes close to strong natural-image pretraining 
at this scale, though the two differ in the scale of 
their pretraining data.

\begin{table*}[t]
\centering
\small
\caption{Licenses, Original and Obtained Source for the ten datasets.}
\label{tab:licenses}
\footnotesize
\setlength{\tabcolsep}{4pt}
\begin{tabular}{@{}llll@{}}
\toprule
Dataset & License & Original source & Obtained from \\
\midrule
SpaceNet 2 & CC BY-SA 4.0 & SpaceNet AWS Open Data & GEO-Bench-2 \cite{geobench2} \\
SpaceNet 3 & CC BY-SA 4.0 & SpaceNet AWS Open Data & Publisher release \\
xView2 (xBD) & CC BY-NC-SA 4.0 & xview2.org & PANGAEA \cite{pangaea} \\
FLAIR 2 & Open Licence 2.0 (Etalab) & IGN France & GEO-Bench-2 \cite{geobench2} \\
Open-Canopy-Small & Open Licence 2.0 (Etalab) & AI4Forest & Size-comparable subset \\
EverWatch & CC0 1.0 & Zenodo & GEO-Bench-2 \cite{geobench2} \\
NZ Cattle & CC BY 4.0 & Zenodo & GEO-Bench-2 \cite{geobench2} \\
m-pv4ger & MIT & Zenodo & GEO-Bench \cite{geobench1} \\
PureForest & Open Licence 2.0 (Etalab) & IGN France & Publisher release \\
HOTOSM Buildings & CC BY 4.0 / ODbL 1.0 & Hugging Face & Publisher release \\
\bottomrule
\end{tabular}
\end{table*}

\subsection{Continual and domain-adaptive pretraining}

Once a strong general-purpose encoder is available, domain specialization does not need to restart pretraining from random initialization. Continual pretraining instead resumes self-supervised training on data from the target domain. This strategy can add domain knowledge at a fraction of the cost of training a foundation model from scratch, but it also risks catastrophic forgetting or degradation of dense features. FINO~\cite{fino}, for example, adapts a visual foundation model while regularizing feature geometry, and related analyses show that unconstrained continued pretraining can affect global and dense representations differently \cite{explora}. Our study uses the original DINOv3 self-supervised objective for a short VHR continuation stage, evaluates the result with the same frozen protocol as the released encoders, and measures where its representation changes using centered kernel alignment \cite{kornblith2019similarity}.

%% file: sec/3_vhr.tex
\section{Very High Resolution: Definition and
Scope}\label{very-high-resolution-definition-and-scope}

We define VHR as RGB optical imagery at a ground sampling distance (GSD)
of 1.5m or finer, and adopt 1.5m as the upper bound for the suite. This
threshold is set by the SPOT 6 and 7 sensors used in the Open-Canopy-Small
dataset, the coarsest platform admitted; all other datasets are finer.
At this resolution, in the terminology of Strahler et
al.~\cite{strahler1986}, the datasets fall in an
H-resolution situation, in which scene elements are larger than the
resolution cell and are individually identifiable, as opposed to the
L-resolution situation of medium-resolution imagery, where elements are
smaller than the cell and are not individually detectable.

The bound is consistent with common usage. The European Space Agency
describes very high resolution as imagery finer than 5m, with the finest
commercial products below
1m.\footnote{ESA, Newcomers Earth Observation Guide, \url{https://business.esa.int/newcomers-earth-observation-guide}.}
The 1.5m threshold sits within this range and excludes coarser
medium-resolution products.

Since most public datasets only use the RGB bands, the study is restricted to 
RGB optical inputs. Nine of the ten datasets
consist of single-date imagery; the exception is xView2, which provides
bi-temporal pre-event and post-event image pairs to support building
damage assessment.

Within this range, the suite spans native imagery at or below 1m GSD,
such as WorldView-3 at 0.3m and aerial surveys at 0.1m, and pansharpened
imagery between 1m and 1.5m, such as SPOT 6/7 at 1.5m, where a
panchromatic band is fused with the multispectral channels. Both native
and pansharpened products are admitted and noted per dataset.

%% file: sec/4_datasets.tex
\section{Benchmark Datasets}\label{benchmark-datasets}

VHR-Bench comprises ten optical-only datasets, all accessed through a
common loading and evaluation interface so that every encoder is
evaluated under the same pipeline. \cref{tab:datasets} lists each dataset with its
sensor, platform, resolution, and region. The suite is balanced across
five task families (\cref{tab:composition}) and
across sensing platforms: four satellite datasets, five aerial, and one mixed
(HOTOSM Buildings), at spatial resolutions from 0.1m to 1.5m.

\cref{tab:licenses} records the license and source
of each dataset. Each license was verified against the original
publisher. Four datasets are reused from GEO-Bench-2,
one from PANGAEA, and one from GEO-Bench. Three are
taken directly from their original publication: SpaceNet 3, PureForest, 
and HOTOSM Buildings. Finally, we create Open-Canopy-Small, a subset of AI4Forest's Open-Canopy to minimize our benchmark's computational footprint; it is published alongside the benchmark (details in \cref{sec:opencanopy-subset}). The splits, tiling, and label encoding follow the source each dataset was taken from. All ten licenses permit research use. Eight allow reuse subject to attribution; EverWatch is
released into the public domain under CC0, and xView2 is released for non-commercial use. Share-alike terms apply
to SpaceNet 2 and 3, to xView2, and to the OpenStreetMap-derived labels
in HOTOSM Buildings. HOTOSM Buildings is the only dataset whose imagery
and labels carry different licenses: imagery from OpenAerialMap is CC BY
4.0 with attribution to contributors of Open Imagery Network, and labels
from OpenStreetMap are ODbL 1.0 with attribution to OpenStreetMap
contributors. The license for m-pv4ger is recorded as MIT in the
GEO-Bench license table and in the publishing repositories.

\begin{figure*}[t]
\centering
\includegraphics[width=\textwidth]{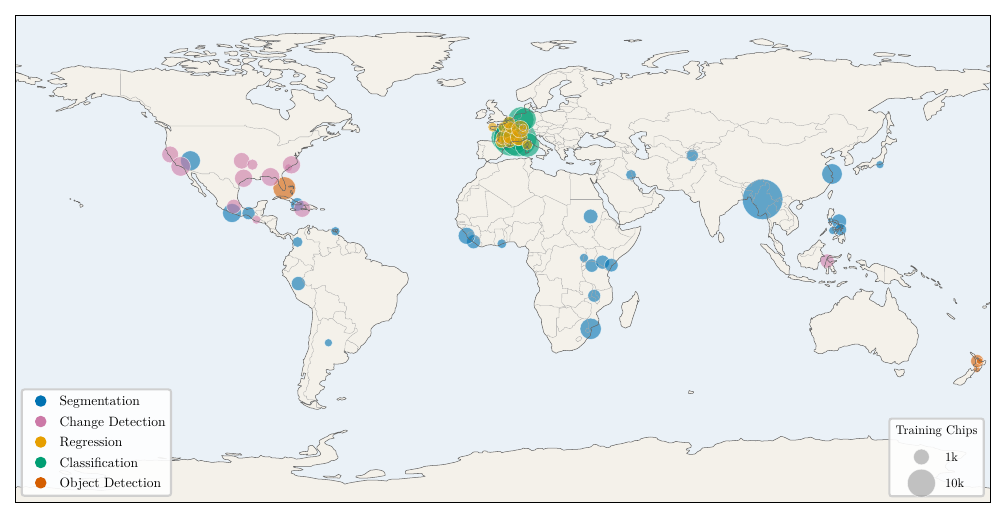}
\caption{Geographic coverage of VHR-Bench. Chip centroids are coloured by task family and bubble size reflects the number of training chips in each area, so small sites stay visible alongside the largest.}
\label{fig:coverage}
\end{figure*}

\cref{tab:composition} summarizes the suite by task
family and application domain. \cref{fig:coverage} displays the geographic coverage of the benchmark. Datasets range geographically from
single-country coverage (France, Germany, New Zealand, the United
States) to genuinely global sets: xView2 spans nineteen disaster events
worldwide, and HOTOSM Buildings covers fifteen countries concentrated in
the Global South, including Myanmar, Mozambique, Mexico, Kenya, and
Liberia. The suite reaches six continents, although the volume is 
concentrated in a few places, and the Global North is slightly over-represented.

\begin{table}[t]
\centering
\footnotesize
\setlength{\tabcolsep}{4pt}
\caption{Task-family and application-domain composition of VHR-Bench; the eight domains are buildings, roads, land cover, building damage, forestry, ecology, agriculture, and energy.}
\label{tab:composition}
\resizebox{\columnwidth}{!}{%
\begin{tabular}{@{}ll>{\raggedright\arraybackslash}p{4.6cm}@{}}
\toprule
Task family & Count & Domain breakdown \\
\midrule
Semantic segmentation & 4 & 2 buildings (SpaceNet 2, HOTOSM Buildings), 1 roads (SpaceNet 3), 1 land cover (FLAIR 2, 13 classes) \\
Change detection & 1 & 1 building damage (xView2) \\
Regression & 1 & 1 forestry / canopy height (Open-Canopy-Small) \\
Object detection & 2 & 1 ecology / wading birds (EverWatch), 1 agriculture / cattle (NZ Cattle) \\
Classification & 2 & 1 energy / PV panels (m-pv4ger, binary), 1 forestry / tree species (PureForest, 13 classes) \\
\bottomrule
\end{tabular}}
\end{table}

\begin{figure*}[t]
\centering
\includegraphics[width=\textwidth]{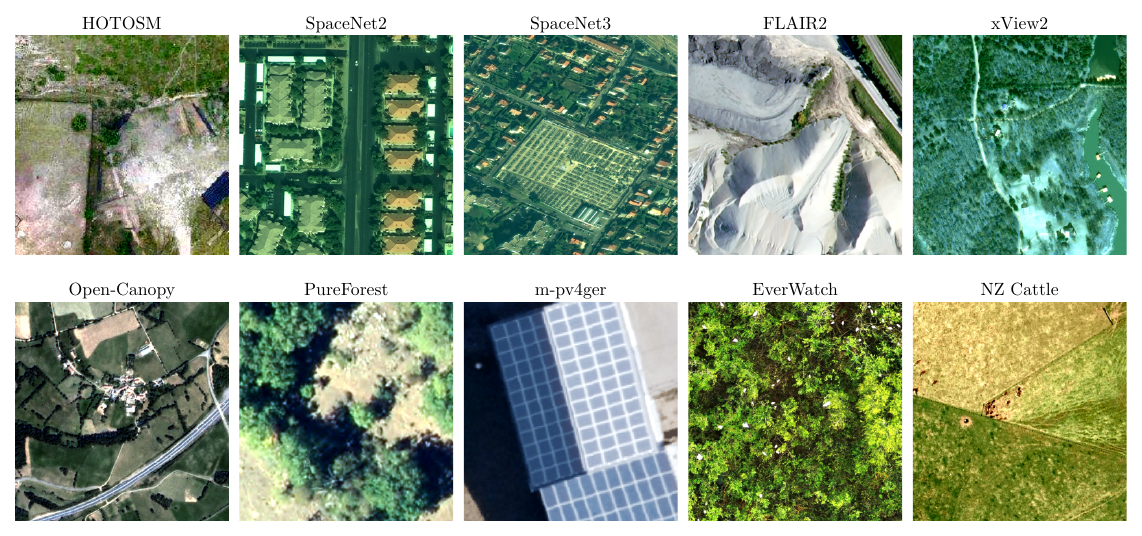}
\caption{One training chip per dataset, drawn through the same loader the encoders use.}
\label{fig:samples}
\end{figure*}

\cref{fig:samples} shows one training chip per
dataset, illustrating how different the scenes and
targets are across the suite. Each task family fixes its decoder,
criterion, and primary metric, described in \cref{tab:protocol}. Every encoder is thus evaluated under the same predetermined configuration. 
The two classification datasets, including the PureForest tree species
dataset~\cite{gaydon2025pureforest}, are class-imbalanced, which
is why macro F1 is their primary metric.

\subsection{Open-Canopy Subset}\label{sec:opencanopy-subset}

The original Open-Canopy dataset weighs approx. 360GB on disk. In an effort to make our benchmark more easily reproducible, we create a subset of the dataset by randomly sampling 10\% of the training and validation patches, while leaving the test set as-is for comparability purposes. The obtained Open-Canopy-Small dataset\footnote{\url{https://huggingface.co/datasets/pierreadorni/Open-Canopy-Small-Comparable}}, weighing 25GB on disk, is used in all our experiments in place of the original Open-Canopy dataset. We also publish a 13GB Open-Canopy-Small Non-Comparable\footnote{\url{https://huggingface.co/datasets/pierreadorni/Open-Canopy-Small}} subset, for which we also reduce the test set to 10\%, although it is not used in this benchmark.

%% file: sec/5_encoders.tex
\section{Encoders}\label{encoders}

Twelve encoders are evaluated, organized into three tiers
defined by pretraining data resolution:

\begin{itemize}
\tightlist
\item
  \textbf{HR-FM}: the pretraining data include high-resolution
  satellite or aerial imagery finer than the Sentinel scale (not exclusively).
\item
  \textbf{CV-FM}: pretrained only on natural or web images, with no
  significant exposure to remote sensing imagery beyond potential occurrences in web data.
\item
  \textbf{MidRes-FM}: pretrained only on medium-resolution
  Sentinel-scale imagery at or above 10m GSD.
\end{itemize}

\cref{tab:encoders} lists the
twelve encoders together with their pretraining data, measured parameter
counts, and forward GFLOPs at each encoder's native input size.

The HR-FM tier contains five encoders; it captures whether an encoder has been exposed to
high-resolution remote sensing imagery during pretraining: its threshold
is therefore less strict than the 1.5m GSD bound used to define the
evaluation data. DINOv3 ViT-L
(SAT)~\cite{dinov3} is the satellite variant of DINOv3,
distilled from the DINOv3 ViT-7B model and pretrained on SAT-493M, a
collection of Maxar RGB tiles at 0.6m GSD.
DOFA~\cite{dofa} is a multimodal foundation model that
uses a wavelength-conditioned dynamic hypernetwork and is
pretrained across five Earth observation sensor modalities.
SkySense~\cite{skysense} is a billion-scale multimodal
remote sensing foundation model pretrained on 21.5 million multimodal
temporal sequences of optical and SAR imagery. It is composed of several modality-specific encoders, which together make the full model billion-scale. We select only the high-resolution optical Swin-Huge backbone, and discard both Sentinel-2 and Sentinel-1 pretrained ViT-L backbones. Clay v1.5~\cite{clay} is a 
masked-autoencoder trained Vision Transformer trained on multi-sensor
inputs including Sentinel-1/2, NAIP, and LINZ (Land Information New Zealand).
EoS-FM~\cite{eosfm} is an ensemble that
fuses the features of multiple lightweight ConvNeXtV2 encoders, each 
pretrained on a different dataset spanning 
Sentinel-1/2 and high-resolution optical imagery.

The CV-FM tier contains four encoders. The DINOv3 ViT-L and ConvNeXt-L
variants~\cite{dinov3} share the DINOv3 self-supervised
objective and are trained on the LVD-1689M web-image collection.
ConvNeXt-XL~\cite{convnext} and ResNet-50~\cite{resnet} are standard
ImageNet-pretrained convolutional baselines. This tier measures how far strong general-purpose visual
features transfer to VHR without any domain-specific pretraining.

The MidRes-FM tier contains three encoders.
TerraMind~\cite{terramind} is a multimodal model
pretrained on several co-registered Earth-observation modalities at
medium resolution, including Sentinel-1 and Sentinel-2. Prithvi-EO
2.0~\cite{prithvi2} is a Vision Transformer pretrained
on Harmonized Landsat and Sentinel-2 imagery at 30m. THOR~\cite{thor} is pretrained
on Sentinel-1/2/3. These encoders have no exposure to
high resolution imagery.

%% file: sec/6_protocol.tex
\section{Evaluation Protocol}\label{evaluation-protocol}

\input{partials/table_encoders.tex}

All encoders are evaluated under one fixed protocol. We freeze the backbone 
and train only an attached task-specific decoder. \cref{tab:protocol} describes the decoder, loss, and primary metric attributed to each task family. Semantic segmentation and change detection report mean intersection over
union (mIoU); change detection uses a Siamese UPerNet that concatenates
the features of the pre-event and post-event images. Canopy height
regression reports root mean squared error (RMSE, lower is better).
Object detection reports mAP@50. Classification reports macro F1, chosen
over accuracy because both classification datasets are class-imbalanced.

The UPerNet and Faster R-CNN decoders both consume a four-level feature pyramid; hierarchical encoders pass their four native stages, whereas features from single-scale vision transformers are extracted from four evenly distributed transformer blocks and resampled to four resolutions using the \verb|Feature2Pyramid| layer from PANGAEA's implementation~\cite{pangaea}.

\begin{table}[t]
\centering
\scriptsize
\setlength{\tabcolsep}{3pt}
\caption{Fixed decoder, loss, and primary metric per task family.}
\label{tab:protocol}
\resizebox{\columnwidth}{!}{%
\begin{tabular}{@{}llll@{}}
\toprule
Task & Decoder & Loss & Metric \\
\midrule
Segmentation & UPerNet & Cross-entropy & mIoU \\
Change detection & Siamese UPerNet & Cross-entropy & mIoU \\
Regression & UPerNet head & MSE & RMSE $\downarrow$ \\
Detection & Faster R-CNN & Cross-entropy + box reg. & mAP@50 \\
Classification & Linear probe & Cross-entropy & Macro F1 \\
\bottomrule
\end{tabular}}
\end{table}

The HR-FM and CV-FM encoders are RGB-native and consume the three RGB
bands directly. The three MidRes-FM encoders natively expect
multispectral input (Prithvi-EO 2.0 with 6, THOR with 10, and TerraMind with 14 channels); the missing
non-RGB channels are filled with zeros, following PANGAEA's band-padding
protocol. The only real spectral signal seen by any encoder is therefore
RGB.

Each encoder is used at its native input size; models supporting a variable image resolution use the fixed size reported in \cref{tab:encoders}. During training, larger dataset patches are randomly cropped to the required dimensions. During validation, we perform sliding-window inference over the whole patch, averaging predictions where they overlap.

We train each decoder with AdamW (learning rate 1e-4, weight decay 0.05,
batch size 8) and a multi-step learning-rate schedule, for a fixed number of 80
 epochs. Every configuration is repeated over three seeds
(42, 1234, 2025) and evaluated both on the full datasets and on a 10\% 
training-data fraction using a stratified subset.

\subsection*{Aggregating scores}
Because the suite combines higher-is-better metrics with one
lower-is-better metric (Open-Canopy-Small RMSE), each per-dataset metric is
mapped to a \textit{closeness to perfection} score in \([0,1]\) before aggregation. For a
dataset \(d\) with raw metric \(m_d\),

\[
c_d =
\begin{cases}
m_d / 100, & \text{\%-valued metrics,}\\
\max\!\bigl(1 - (m_d / r)^2,\ \epsilon\bigr),
& \text{RMSE,}
\end{cases}
\]

where \(\epsilon = 10^{-6}\) ensures that the logarithm remains defined,
and \(r\) is the population standard deviation of canopy height over
the Open-Canopy-Small test split\footnote{\(r = 85.907\), computed over the
13,675 test tiles.}. Since \(r\) is computed using the same
samples and weighting as the RMSE, the unclipped Open-Canopy-Small score is
the coefficient of determination \(R^2\); predictions worse than the
mean-target baseline are assigned \(\epsilon\). The ten scores are
combined per seed using their geometric mean,

\[
g = \left(\prod_{d=1}^{10} c_d\right)^{1/10}
  = \exp\!\left(\frac{1}{10}\sum_{d=1}^{10} \ln c_d\right),
\]

and the reported aggregate is the mean of \(g\) over the three seeds. 
For readability, we report the aggregate score as \(G=100g\), placing it on a \(0\)–\(100\) scale.
Anchoring each score to a perfect result makes the aggregate independent
of the set of encoders being compared: adding or removing an encoder
does not rescale the scores of the others. Moreover, compared with an
arithmetic mean, the geometric mean penalizes low performance on any
individual dataset more strongly. The aggregate is
computed at each label fraction, and encoders are ranked overall by the
mean of their aggregate at the two fractions, so the low-label and
full-label regimes count equally, while the per-fraction leaders are
reported alongside.

\subsection*{Reproducibility}

We release a config-driven evaluation framework
built as an extended fork of the PANGAEA codebase, intended for seamless reproducibility and upstream contribution. We content-hash and mirrored the datasets and the encoder checkpoints on Hugging Face, so the exact data and weights can be verified. The evaluation code, configurations, and dataset manifests will be released\footnote{upon acceptance.} so the full suite can be re-run. The software environment is pinned and provided in the repository: Python 3.13 with PyTorch 2.12.0 (CUDA 13.0), torchvision 0.27.0, timm 1.0.27, and torchmetrics 1.9.0.

%% file: partials/table_encoders.tex
\begin{table*}[t]
\centering
\small
\setlength{\tabcolsep}{3pt}
\caption[Evaluated encoders]{Evaluated encoders. A dash in the Patch or Tokens column means the quantity does not apply, as for the convolutional encoders and for SkySense, whose windowed attention makes the token count vary by stage.}
\label{tab:encoders}
\begin{tabular}{@{}rllrrrrrl@{}}
\toprule
\# & Tier & Encoder & Params (M) & GFLOPs & Input (px) & Patch & Tokens & Pretraining data \\
\midrule
1 & HR-FM & Clay v1.5 ViT-L \cite{clay} & 311 & 723.0 & 256 & 8 & 1024 & Sentinel-1/2, NAIP, LINZ \\
2 &  & DINOv3 ViT-L (SAT) \cite{dinov3} & 303 & 164.7 & 256 & 16 & 256 & SAT-493M (Maxar, 0.6m) \\
3 &  & DOFA ViT-B \cite{dofa} & 111 & 35.4 & 224 & 16 & 196 & Multi-sensor (sat + aerial) \\
4 &  & EoS-FM \cite{eosfm} & 78 & 206.9 & 256 & -- & -- & Sentinel-1/2 and HR optical \\
5 &  & SkySense SwinH \cite{skysense} & 655 & 301.6 & 256 & 4 & -- & WorldView 3/4, Sentinel-1/2 \\
6 & CV-FM & ConvNeXt-XL \cite{convnext} & 348 & 159.1 & 256 & -- & -- & ImageNet-22k to 1k \\
7 &  & DINOv3 ConvNeXt-L \cite{dinov3} & 196 & 89.8 & 256 & -- & -- & LVD-1689M (web) \\
8 &  & DINOv3 ViT-L (LVD) \cite{dinov3} & 303 & 164.7 & 256 & 16 & 256 & LVD-1689M (web) \\
9 &  & ResNet-50 \cite{resnet} & 24 & 8.2 & 224 & -- & -- & ImageNet-1k \\
10 & MidRes-FM & Prithvi-EO 2.0 600M \cite{prithvi2} & 631 & 335.0 & 224 & 14 & 256 & HLS (Landsat + S2), 30m \\
11 &  & THOR v1 Base \cite{thor} & 94 & 44.2 & 224 & 16 & 196 & Sentinel-1/2/3 \\
12 &  & TerraMind v1 L \cite{terramind} & 306 & 253.3 & 224 & 16 & 196 & Sentinel-1/2, DEM, LULC \\
\bottomrule
\end{tabular}
\end{table*}

%% file: sec/7_continued_pretraining.tex
\section{Continued VHR Pretraining}
\label{sec:continued-pretraining}

\begin{table*}[t]
\centering
\footnotesize
\setlength{\tabcolsep}{3pt}
\caption{VHR continued-pretraining corpus. Planet tiles are open disaster
releases from Venezuela, the Philippines, and Jamaica.}
\label{tab:pretrain-data}
\begin{tabular}{@{}lrlllll@{}}
\toprule
Source & Tiles & Platform & Sensor & GSD & License & Obtained from \\
\midrule
fMoW \cite{fmow} & 500,008 & Satellite & QuickBird, GeoEye, WorldView & 0.3--0.7m & fMoW Challenge (NC) & AWS Open Data \\
Maxar Open Data \cite{maxar_open_data} & 120,000 & Satellite & WorldView & 0.3m & CC BY-NC 4.0 & Maxar STAC \\
OpenAerialMap \cite{openaerialmap} & 100,000 & Drone/aircraft & Mixed & 0.05--1m & CC BY 4.0 & OAM STAC \\
NAIP \cite{naip} & 100,000 & Aerial & Aerial RGB & 1.25m & Public domain & SatlasPretrain~\cite{satlas} \\
RAMP \cite{ramp} & 95,333 & Satellite & WorldView, GeoEye & 0.3--0.5m & CC BY-NC 4.0 & Source Cooperative \\
WorldStrat \cite{worldstrat} & 62,822 & Satellite & SPOT 6/7 & 1.5m & CC BY-NC 4.0 & Zenodo~\cite{worldstrat} \\
Planet Disaster Data \cite{planet_disaster_data} & 40,000 & Satellite & SkySat, Pelican & 0.5m & CC BY-NC 4.0 & Source Cooperative \\
\midrule
Total & 1,018,163 & & & & & \\
\bottomrule
\end{tabular}
\end{table*}

\begin{figure*}[t]
\centering
\includegraphics[width=\textwidth]{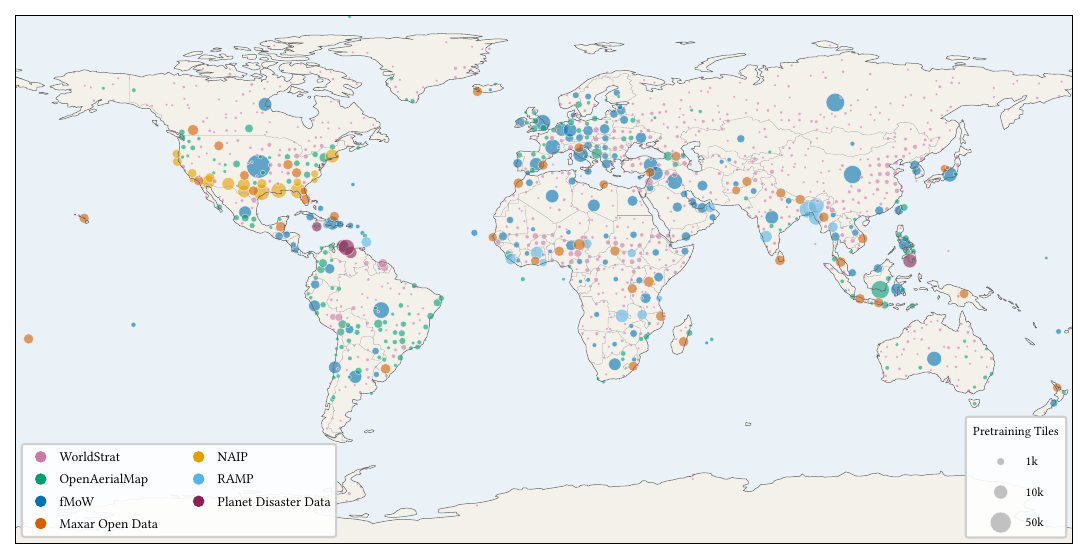}
\caption{Geographic coverage of the continued-pretraining corpus. Tile
footprints are aggregated on a coarse grid; bubble area is proportional to the
number of tiles and colour identifies the dominant source in each cell.}
\label{fig:pretrain-coverage}
\end{figure*}

The released-encoder comparison (detailed results in \cref{sec:results}) 
identifies DINOv3 ViT-L (LVD) as the
strongest encoder at the full label fraction and SkySense as the strongest when
labels are scarce. It also shows a task-level distinction between the two matched
DINOv3 ViT-L variants: the web-pretrained model is stronger on semantic tasks,
whereas the satellite-pretrained model retains an advantage on canopy-height
regression and object detection. This motivates a benchmark-guided intervention:
starting from the web model, we ask whether a short self-supervised continuation on
VHR imagery can add domain knowledge without discarding its broadly transferable
features.

\subsection{VHR pretraining corpus}
\label{sec:vhr-pretraining-corpus}

We assemble 1,018,163 non-overlapping $256\times256$ RGB tiles from seven
sources (\cref{tab:pretrain-data}). Their GSDs range from 0.05 to 1.5m and
include satellite, aerial, and drone imagery over six continents. The largest
source, fMoW~\cite{fmow}, is capped at approximately 500,000 tiles, and NAIP is
capped at 100,000 tiles drawn from the SatlasPretrain pool~\cite{satlas}, so
neither archive dominates the mixture. Sources are sampled in proportion to the
tile counts in \cref{tab:pretrain-data}. The resulting WebDataset shards include
source, license, GSD, and tile-coordinate metadata. Because the most restrictive
source terms apply to the combined collection, the corpus is released under
CC BY-NC-SA 4.0.

\subsubsection*{Separation from VHR-Bench}
The ten benchmark datasets are not included as pretraining sources. We additionally
exclude known intersections with the broader archives used to construct the
corpus. Maxar disaster events appearing in xView2 are removed from Maxar Open
Data; the RAMP Shanghai and Paris subsets are removed because SpaceNet 2 and 3
evaluate imagery from those cities; and OpenAerialMap mosaics used by HOTOSM
Buildings are excluded. The other benchmark datasets originate from national
mapping programs or acquisitions outside the corpus sources. These exclusions are
performed using the available source and geographic metadata. Residual geographic
overlap is still possible where an archive lacks complete tile-level provenance,
but no benchmark images or labels are intentionally used during continued
pretraining.

\subsection{Continued-pretraining method}
\label{sec:continued-method}

We initialize the student and exponential-moving-average teacher from the released
DINOv3 ViT-L LVD-1689M backbone. The released checkpoint does not contain the
self-supervised projection heads, so the DINO and iBOT heads are reinitialized. We
continue the original DINOv3 objective,

\begin{equation}
\mathcal{L}_{\mathrm{VHR}} =
\mathcal{L}_{\mathrm{DINO}} +
\mathcal{L}_{\mathrm{iBOT}} +
0.1\,\mathcal{L}_{\mathrm{KoLeo}},
\end{equation}

which combines class-token self-distillation, masked-patch prediction, and an
embedding-spread regularizer~\cite{dinov3}. Task labels were not used in this objective.
The reported model is the unconstrained continuation; feature-preservation anchors
were tested separately but did not improve downstream transfer at the evaluated
setting.

Randomly initialized heads can send destructive gradients into an already useful
backbone. We therefore freeze the transformer for the first 1,000 steps while the
heads stabilize, leaving only the patch-embedding stem trainable so it can adjust
to VHR pixel statistics. The transformer then unfreezes through a 500-step linear
warmup to the operating learning rate. \Cref{tab:continued-config} gives the full
configuration. Tiles are streamed with replacement; at an effective batch of 512,
the first-pass checkpoint processes approximately one million tile draws and covers
about 63\% of the distinct corpus.

\begin{table}[t]
\centering
\footnotesize
\caption{Continued-pretraining configuration.}
\label{tab:continued-config}
\begin{tabular}{@{}ll@{}}
\toprule
Setting & Value \\
\midrule
Initialization & DINOv3 ViT-L (LVD-1689M) \\
Input & $256\times256$ RGB \\
Effective batch & 512 \\
Objective & DINO + iBOT + 0.1 KoLeo \\
Schedule & Freeze 1,000; warm up 500 steps \\
Backbone learning rate & $5\times10^{-5}$ \\
Reported checkpoint & First pass ($\sim$2,000 steps) \\
Hardware & 2$\times$A100 80GB \\
\bottomrule
\end{tabular}
\end{table}

The operating learning rate is selected without VHR-Bench labels, using
self-supervised loss stability, ImageNet nearest-neighbour retention, and feature
standard deviation. Training-length checkpoints at approximately 2,000, 4,000, and
6,000 steps are then compared on the validation set of a four-task development subset at 10\% labels.
Classification degrades after the first pass while dense-task performance is flat
or changes only slightly, so we report the first-pass checkpoint. The complete
ten-dataset test suite is evaluated with the frozen-backbone protocol of
\cref{evaluation-protocol}; it is not used to update the continued model.

For direct comparison with the released encoders, the primary checkpoint is
evaluated over the same three decoder seeds. We also repeat continued pretraining
with seeds 42, 1234, and 2025 at the same data and compute budget, holding the
decoder seed fixed, to measure run-to-run stability separately from decoder
variability. We refer to the resulting encoder as DINOv3 ViT-L
(LVD$\rightarrow$HR), or the continued model.

%% file: sec/8_results.tex
\section{Results}\label{sec:results}

The full-dataset benchmark ran for 1480 GPU-h on H100 GPUs, averaging 38h to test an encoder on the full suite once. The 10\% data split took 309 GPU-h, averaging $\sim$8h for a single encoder to run once on the full suite.

\input{partials/table_ranking.tex}

\subsection{Released-encoder benchmark}

\begin{figure*}[t]
\centering
\includegraphics[width=\textwidth]{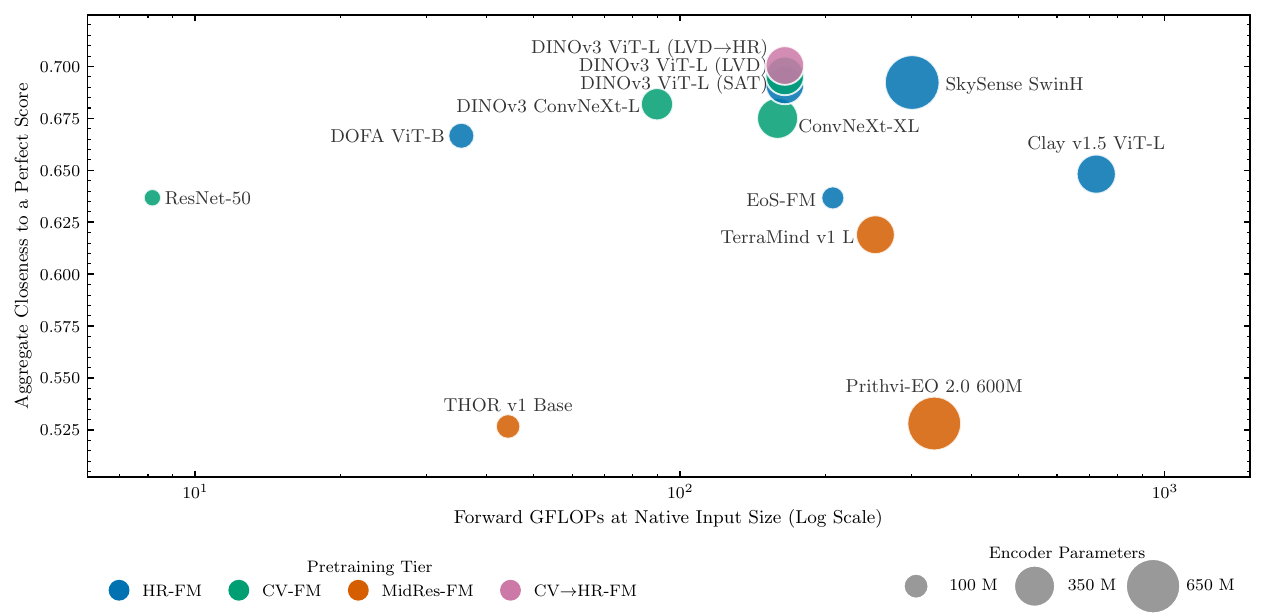}
\caption{Aggregate closeness versus forward GFLOPs; marker size scales with parameter count, colour with the pretraining tier. Patch size sets the token count: Clay v1.5 (patch 8) carries 1024 tokens and 723 GFLOPs against DINOv3 ViT-L (patch 16) at 256 tokens and 164.7 GFLOPs.}
\label{fig:compute}
\end{figure*}

\begin{figure}[t]
\centering
\includegraphics[width=\columnwidth]{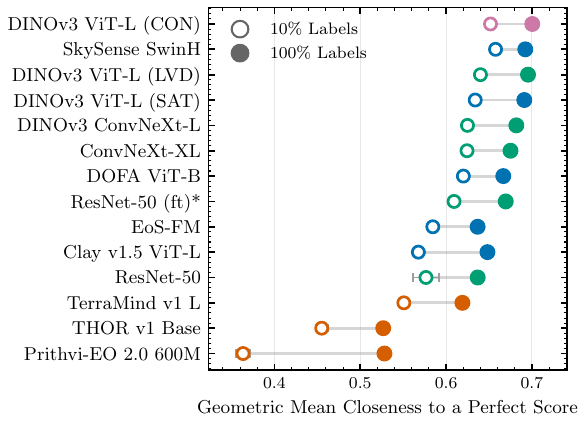}
\caption{Overall geometric-mean closeness at both training fractions, coloured by pretraining tier. Error bars are one standard deviation over three seeds. Encoders are ordered by the mean of the two fractions. (ft)* means fully fine-tuned. CON designates our continual pretraining experiment (LVD$\rightarrow$HR).}
\label{fig:ranking}
\end{figure}

\cref{tab:ranking} and \cref{fig:ranking} report the aggregate score for
every encoder at both training-data fractions (10\% and 100\%). Besides our continued model,
DINOv3 ViT-L (LVD) obtains the highest full-fraction average
(69.5), followed by SkySense (69.2), and DINOv3 ViT-L (SAT) (69.1).
These three top scores are separated by a few tenths of a point.
At 10\% labels, SkySense leads more clearly, at 65.8 points against 64.0 for DINOv3 ViT-L (LVD) and 63.4 for its SAT counterpart. The first six released
positions are divided equally between CV-FM and HR-FM encoders, whereas the
three MidRes-FM encoders occupy the last three positions. Every encoder improves
from the reduced to the full label fraction, and the ordering is largely preserved.

The two DINOv3 ViT-L encoders provide a matched comparison in which the architecture, parameter count, input size, and evaluation protocol are held fixed while the pretraining data change. The web-pretrained LVD variant obtains a slightly higher aggregate score at both label fractions: 64.0 versus 63.4 at 10\%, and 69.5 versus 69.1 at 100\%. The aggregate proximity nevertheless masks a consistent task-level difference. At the full fraction, the LVD variant performs better on all four semantic-segmentation datasets, change detection, and both classification datasets, whereas the SAT variant performs better on canopy-height regression and both object-detection datasets. At the 10\% fraction, LVD leads on eight datasets, while SAT leads on Open-Canopy-Small and NZ Cattle. In particular, SAT records the lowest Open-Canopy-Small RMSE at both fractions: 38.2 compared with 40.5 for the next-best encoder at 10\%, and 35.1 compared with 36.6 at 100\%.

\subsection{Effect of continued VHR pretraining}

The continued model obtains the highest observed full-fraction aggregate,
70.0, compared with 69.5 for its released initialization, 69.2 for SkySense, and
69.1 for DINOv3 ViT-L (SAT). At 10\% labels it reaches 65.2, ranking second behind
SkySense at 65.8 and ahead of LVD at 64.0 and SAT at 63.4. Its two-fraction mean
is 67.6, effectively tied with SkySense at 67.5 given the run-to-run spread. Since
continued pretraining changes only the weights, its 303M parameters and 164.7
forward GFLOPs are identical to those of the two released DINOv3 ViT-L variants.

The aggregate gain over the initialization is larger when labels are scarce
(+1.2 points) than at the full fraction (+0.5). At 10\% labels, the largest
per-dataset change is NZ Cattle detection, which rises from 59.1 to 67.0
mAP@50. At full labels, the clearest gains occur on FLAIR 2 land-cover
segmentation (54.3 to 55.4 mIoU) and PureForest classification (71.5 to 72.6
macro F1). Performance on the building, road, and change-detection datasets is
largely preserved. The continuation does not uniformly inherit the SAT model's
geometric strengths: at full labels SAT remains ahead of LVD on Open-Canopy-Small and both detection datasets, while SkySense records the strongest detection results.

Tables~\ref{tab:results10} and~\ref{tab:results100} give the complete
per-dataset results. The continued model is best or tied-best among frozen
encoders on four datasets at 10\% labels and five at full labels. Across the suite,
the largest variance remains on NZ Cattle and the smallest on the saturated
m-pv4ger classification task. The fully fine-tuned ResNet-50 reaches 60.9 and
67.0 in aggregate, improving over its frozen counterpart but remaining below the
leading frozen encoders. It obtains the highest m-pv4ger score at both fractions
and the highest full-fraction EverWatch score.

\input{partials/table_results_10pct.tex}

\subsection{Pretraining repeatability and feature retention}

The primary ranking treats the continued checkpoint like every released
checkpoint and reports variation over three decoder seeds. To separate this from
continued-pretraining variability, we repeat the VHR continuation three times at a
matched data and compute budget and evaluate each run with decoder seed 42
(\cref{tab:continued-repeatability}). The full-fraction aggregate remains 70.0
with a standard deviation of 0.1. At 10\% labels it averages 64.8 with a standard
deviation of 0.2, preserving its position between SkySense and the released LVD
model. The exact two-fraction ordering of the continued model and SkySense is
therefore not treated as a meaningful separation; the stable result is that both
belong to the leading group and that continuation improves on its initialization.

\begin{table}[t]
\centering
\footnotesize
\caption{Aggregate closeness of three independent continued-pretraining runs.
The decoder seed is fixed at 42 to isolate pretraining variability.}
\label{tab:continued-repeatability}
\begin{tabular}{@{}lrrrr@{}}
\toprule
Fraction & Seed 42 & Seed 1234 & Seed 2025 & Mean $\pm$ std \\
\midrule
10\% & 65.1 & 64.7 & 64.7 & 64.8$_{\pm0.2}$ \\
100\% & 69.9 & 69.9 & 70.1 & 70.0$_{\pm0.1}$ \\
\bottomrule
\end{tabular}
\end{table}

Centered kernel alignment (CKA) between the continued and released encoders shows
where specialization occurs. Patch-token similarity remains above 0.99 in the
first half of the backbone, falls to 0.922 at layer 17, and reaches 0.789 at the
last probed layer (\cref{tab:continued-cka}). ImageNet nearest-neighbour accuracy
is preserved, moving from 71.9\% before continuation to 72.1\% at the reported
checkpoint. The continued model therefore retains the early and general-purpose
representation while adapting primarily in its deepest layers.

\begin{table}[t]
\centering
\footnotesize
\caption{Patch-token CKA between the continued and released DINOv3 ViT-L
encoders; mean and standard deviation over three continued-pretraining runs.}
\label{tab:continued-cka}
\begin{tabular}{@{}lrrrr@{}}
\toprule
Layer & 5 & 11 & 17 & 23 \\
\midrule
CKA & 0.998$_{\pm0.000}$ & 0.996$_{\pm0.000}$ & 0.922$_{\pm0.012}$ & 0.789$_{\pm0.016}$ \\
\bottomrule
\end{tabular}
\end{table}

\subsection{Ranking sensitivity and compute}

\input{partials/table_results_100pct.tex}

We assess the sensitivity of the aggregate ranking by recomputing it after removing each dataset in turn. The full-suite leader remains unchanged in nineteen of the twenty leave-one-out comparisons across the two label fractions (\cref{tab:loo}). The only change occurs when NZ Cattle is removed at 10\% labels, in which case the continued model replaces SkySense as the leader. The full-fraction lead of the continued model is unchanged under every single-dataset removal. 

\Cref{fig:compute} plots aggregate score against encoder computation. A top-tier
result does not require the largest model: the three DINOv3 ViT-L checkpoints use
roughly half the parameters and compute of SkySense, while the most expensive
encoder, Clay, remains mid-ranked. Continued VHR pretraining moves DINOv3 upward
on the accuracy axis without moving it on the inference-cost axis.

\input{partials/table_loo.tex}

%% file: partials/table_ranking.tex
\begin{table}[t]
\centering
\small
\setlength{\tabcolsep}{2pt}
\caption[Overall ranking]{Overall ranking by geometric-mean closeness to a perfect score. The mean over three seeds and standard deviation are reported, with the best result per fraction in bold. CON designates our continual pretraining experiment (LVD$\rightarrow$HR). Italics and $^{*}$ denote fully fine-tuned models; otherwise, backbones are frozen. Closeness is scaled by 100.}
\label{tab:ranking}
\begin{tabular}{@{}rllrrr@{}}
\toprule
\# & Tier & Encoder & 10\% & 100\% & Mean \\
\midrule
1 & CON & DINOv3 ViT-L (CON) & 65.2$_{\pm0.3}$ & \textbf{70.0}$_{\pm0.2}$ & 67.6 \\
2 & HR & SkySense SwinH & \textbf{65.8}$_{\pm0.4}$ & 69.2$_{\pm0.1}$ & 67.5 \\
3 & CV & DINOv3 ViT-L (LVD) & 64.0$_{\pm0.3}$ & 69.5$_{\pm0.1}$ & 66.8 \\
4 & HR & DINOv3 ViT-L (SAT) & 63.4$_{\pm0.2}$ & 69.1$_{\pm0.3}$ & 66.3 \\
5 & CV & DINOv3 ConvNeXt-L & 62.5$_{\pm0.1}$ & 68.2$_{\pm0.2}$ & 65.3 \\
6 & CV & ConvNeXt-XL & 62.4$_{\pm0.0}$ & 67.5$_{\pm0.2}$ & 65.0 \\
7 & HR & DOFA ViT-B & 62.0$_{\pm0.3}$ & 66.7$_{\pm0.2}$ & 64.3 \\
8 & HR & EoS-FM & 58.5$_{\pm0.2}$ & 63.7$_{\pm0.5}$ & 61.1 \\
9 & HR & Clay v1.5 ViT-L & 56.8$_{\pm0.5}$ & 64.8$_{\pm0.2}$ & 60.8 \\
10 & CV & ResNet-50 & 57.7$_{\pm1.5}$ & 63.7$_{\pm0.1}$ & 60.7 \\
11 & MidRes & TerraMind v1 L & 55.1$_{\pm0.5}$ & 61.9$_{\pm0.2}$ & 58.5 \\
12 & MidRes & THOR v1 Base & 45.5$_{\pm0.7}$ & 52.7$_{\pm0.2}$ & 49.1 \\
13 & MidRes & Prithvi-EO 2.0 600M & 36.4$_{\pm0.8}$ & 52.8$_{\pm0.3}$ & 44.6 \\
\midrule
-- & \textit{Baseline} & \textit{ResNet-50 (ft)}$^{*}$ & \textit{60.9}$_{\pm0.1}$ & \textit{67.0}$_{\pm0.3}$ & \textit{63.9} \\
\bottomrule
\end{tabular}
\end{table}

%% file: partials/table_results_10pct.tex
\begin{table*}[t]
\centering
\scriptsize
\setlength{\tabcolsep}{3pt}
\caption[Per-dataset metric at 10\% labels]{Primary metric per dataset at 10 percent training data, frozen backbone except indicated. Segmentation and change detection report mIoU, classification macro F1, detection mAP@50, Open-Canopy-Small RMSE in decimetres (lower is better). Mean over three decoder seeds, subscript is the standard deviation; best frozen-encoder mean in bold. The fully fine-tuned baseline is not counted when bolding. Datasets: HOT: HOTOSM Buildings; SN2, SN3: SpaceNet 2, 3; FLR: FLAIR2; xV2: xView2; OpC: Open-Canopy-Small; PF: PureForest; PV: m-pv4ger; EvW: EverWatch; NZC: NZ Cattle.}
\label{tab:results10}
\resizebox{\textwidth}{!}{%
\begin{tabular}{@{}llrrrrrrrrrr@{}}
\toprule
Tier & Encoder & HOT & SN2 & SN3 & FLR & xV2 & OpC & PF & PV & EvW & NZC \\
\midrule
CV$\rightarrow$VHR & DINOv3 ViT-L (LVD$\rightarrow$HR) & \textbf{66.5}$_{\pm0.4}$ & \textbf{87.6}$_{\pm0.0}$ & 69.7$_{\pm0.4}$ & \textbf{53.9}$_{\pm0.6}$ & \textbf{57.4}$_{\pm0.6}$ & 39.8$_{\pm0.1}$ & 66.2$_{\pm0.4}$ & 93.4$_{\pm0.2}$ & 33.9$_{\pm0.8}$ & 67.0$_{\pm2.8}$ \\
\midrule
HR-FM & Clay v1.5 ViT-L & 61.2$_{\pm0.8}$ & 79.9$_{\pm0.3}$ & 58.8$_{\pm0.5}$ & 43.9$_{\pm0.1}$ & 51.8$_{\pm0.6}$ & 42.8$_{\pm0.1}$ & 45.7$_{\pm0.3}$ & 82.2$_{\pm0.8}$ & 28.3$_{\pm1.5}$ & 66.8$_{\pm1.1}$ \\
 & DINOv3 ViT-L (SAT) & 65.4$_{\pm0.7}$ & 85.3$_{\pm0.3}$ & 68.6$_{\pm0.4}$ & 51.2$_{\pm0.3}$ & 55.9$_{\pm0.3}$ & \textbf{38.2}$_{\pm0.1}$ & 63.0$_{\pm1.5}$ & 90.2$_{\pm0.2}$ & 31.9$_{\pm0.7}$ & 66.0$_{\pm1.5}$ \\
 & DOFA ViT-B & 63.1$_{\pm1.0}$ & 83.8$_{\pm0.2}$ & 65.5$_{\pm0.5}$ & 48.9$_{\pm0.3}$ & 53.7$_{\pm0.3}$ & 42.7$_{\pm0.1}$ & 61.4$_{\pm1.6}$ & 92.3$_{\pm0.9}$ & 32.2$_{\pm0.5}$ & 67.6$_{\pm0.8}$ \\
 & EoS-FM & 61.9$_{\pm0.5}$ & 82.9$_{\pm0.1}$ & 66.8$_{\pm0.7}$ & 41.6$_{\pm0.8}$ & 53.2$_{\pm0.3}$ & 43.3$_{\pm0.1}$ & 49.1$_{\pm1.2}$ & 89.4$_{\pm0.7}$ & 27.3$_{\pm0.6}$ & 68.8$_{\pm1.2}$ \\
 & SkySense SwinH & 65.9$_{\pm0.3}$ & 86.5$_{\pm0.1}$ & 67.5$_{\pm0.5}$ & 52.0$_{\pm0.2}$ & 57.1$_{\pm0.6}$ & 40.5$_{\pm0.1}$ & \textbf{67.1}$_{\pm2.2}$ & 93.6$_{\pm0.7}$ & \textbf{35.9}$_{\pm0.9}$ & \textbf{75.6}$_{\pm0.7}$ \\
CV-FM & ConvNeXt-XL & 63.8$_{\pm0.3}$ & 84.3$_{\pm0.3}$ & 65.8$_{\pm0.4}$ & 48.3$_{\pm0.2}$ & 53.7$_{\pm0.5}$ & 42.7$_{\pm0.1}$ & 62.1$_{\pm2.6}$ & \textbf{93.8}$_{\pm1.0}$ & 34.6$_{\pm0.4}$ & 64.7$_{\pm2.6}$ \\
 & DINOv3 ConvNeXt-L & 65.8$_{\pm0.3}$ & 86.3$_{\pm0.1}$ & 66.7$_{\pm0.1}$ & 51.9$_{\pm0.1}$ & 54.6$_{\pm0.5}$ & 42.9$_{\pm0.1}$ & 65.4$_{\pm0.5}$ & 93.7$_{\pm0.3}$ & 30.8$_{\pm0.6}$ & 59.8$_{\pm0.3}$ \\
 & DINOv3 ViT-L (LVD) & \textbf{66.5}$_{\pm0.6}$ & \textbf{87.6}$_{\pm0.1}$ & \textbf{69.8}$_{\pm0.3}$ & 53.7$_{\pm0.1}$ & 56.9$_{\pm0.3}$ & 40.7$_{\pm0.1}$ & 65.9$_{\pm0.9}$ & 92.6$_{\pm0.6}$ & 33.3$_{\pm1.1}$ & 59.1$_{\pm2.0}$ \\
 & ResNet-50 & 58.8$_{\pm0.4}$ & 80.7$_{\pm0.5}$ & 64.5$_{\pm0.1}$ & 44.6$_{\pm0.4}$ & 49.2$_{\pm0.5}$ & 53.4$_{\pm6.2}$ & 54.7$_{\pm0.7}$ & 91.0$_{\pm0.6}$ & 28.4$_{\pm1.1}$ & 71.2$_{\pm1.0}$ \\
MidRes-FM & Prithvi-EO 2.0 600M & 57.6$_{\pm1.1}$ & 74.9$_{\pm0.5}$ & 57.4$_{\pm1.3}$ & 36.1$_{\pm0.3}$ & 36.6$_{\pm0.2}$ & 51.9$_{\pm0.3}$ & 40.9$_{\pm1.4}$ & 88.6$_{\pm0.5}$ & 8.8$_{\pm0.3}$ & 6.2$_{\pm1.3}$ \\
 & THOR v1 Base & 54.0$_{\pm1.3}$ & 72.9$_{\pm0.3}$ & 57.3$_{\pm1.1}$ & 29.9$_{\pm1.2}$ & 40.6$_{\pm0.3}$ & 52.4$_{\pm0.1}$ & 37.5$_{\pm1.1}$ & 86.8$_{\pm0.6}$ & 19.6$_{\pm1.6}$ & 34.9$_{\pm0.7}$ \\
 & TerraMind v1 L & 56.8$_{\pm2.0}$ & 79.9$_{\pm0.1}$ & 61.6$_{\pm0.1}$ & 40.6$_{\pm0.1}$ & 48.4$_{\pm0.2}$ & 45.6$_{\pm0.0}$ & 48.4$_{\pm2.6}$ & 87.9$_{\pm1.9}$ & 26.3$_{\pm0.9}$ & 58.3$_{\pm1.0}$ \\
\midrule
Baseline & ResNet-50 (ft)$^{*}$ & 61.9$_{\pm0.6}$ & 81.7$_{\pm0.8}$ & 65.4$_{\pm0.1}$ & 44.9$_{\pm0.5}$ & 52.1$_{\pm0.6}$ & 47.2$_{\pm2.9}$ & 56.4$_{\pm1.4}$ & 94.9$_{\pm0.9}$ & 33.8$_{\pm1.2}$ & 72.4$_{\pm0.7}$ \\
\bottomrule
\end{tabular}
}
\end{table*}

%% file: partials/table_results_100pct.tex
\begin{table*}[t]
\centering
\scriptsize
\setlength{\tabcolsep}{3pt}
\caption[Per-dataset metric at 100\% labels]{Primary metric per dataset at 100 percent training data, frozen backbone except indicated. Metrics and dataset abbreviations as in Table~\ref{tab:results10}. Mean over three decoder seeds, subscript is the standard deviation; best frozen-encoder mean in bold. The fully fine-tuned baseline is not counted when bolding.}
\label{tab:results100}
\resizebox{\textwidth}{!}{%
\begin{tabular}{@{}llrrrrrrrrrr@{}}
\toprule
Tier & Encoder & HOT & SN2 & SN3 & FLR & xV2 & OpC & PF & PV & EvW & NZC \\
\midrule
CV$\rightarrow$VHR & DINOv3 ViT-L (LVD$\rightarrow$HR) & \textbf{66.7}$_{\pm0.2}$ & \textbf{88.9}$_{\pm0.0}$ & \textbf{73.2}$_{\pm0.1}$ & \textbf{55.4}$_{\pm0.1}$ & 61.4$_{\pm0.3}$ & 36.1$_{\pm0.1}$ & \textbf{72.6}$_{\pm0.5}$ & 95.3$_{\pm0.1}$ & 43.2$_{\pm0.5}$ & 78.1$_{\pm0.4}$ \\
\midrule
HR-FM & Clay v1.5 ViT-L & 61.6$_{\pm0.5}$ & 84.9$_{\pm0.1}$ & 67.8$_{\pm0.1}$ & 45.9$_{\pm0.1}$ & 58.5$_{\pm0.3}$ & 37.6$_{\pm0.0}$ & 57.5$_{\pm0.0}$ & 89.1$_{\pm0.1}$ & 42.8$_{\pm0.6}$ & 77.7$_{\pm0.7}$ \\
 & DINOv3 ViT-L (SAT) & 64.8$_{\pm0.4}$ & 88.0$_{\pm0.1}$ & 72.5$_{\pm0.1}$ & 52.4$_{\pm0.2}$ & 60.9$_{\pm0.2}$ & \textbf{35.1}$_{\pm0.1}$ & 70.7$_{\pm1.4}$ & 92.1$_{\pm0.2}$ & 43.5$_{\pm0.9}$ & 79.8$_{\pm0.4}$ \\
 & DOFA ViT-B & 62.8$_{\pm1.2}$ & 86.8$_{\pm0.1}$ & 70.7$_{\pm0.1}$ & 49.9$_{\pm0.2}$ & 59.3$_{\pm0.1}$ & 38.7$_{\pm0.0}$ & 65.7$_{\pm0.7}$ & 93.8$_{\pm0.1}$ & 40.1$_{\pm0.2}$ & 77.3$_{\pm0.3}$ \\
 & EoS-FM & 63.0$_{\pm0.8}$ & 85.3$_{\pm0.0}$ & 71.1$_{\pm0.1}$ & 44.3$_{\pm0.8}$ & 56.9$_{\pm1.6}$ & 39.2$_{\pm0.1}$ & 54.9$_{\pm0.9}$ & 90.9$_{\pm0.7}$ & 36.7$_{\pm1.1}$ & 78.6$_{\pm1.3}$ \\
 & SkySense SwinH & 66.1$_{\pm0.4}$ & 88.4$_{\pm0.0}$ & 72.4$_{\pm0.1}$ & 53.1$_{\pm0.2}$ & \textbf{61.6}$_{\pm0.1}$ & 36.6$_{\pm0.2}$ & 66.2$_{\pm0.4}$ & 94.3$_{\pm0.7}$ & \textbf{43.9}$_{\pm0.4}$ & \textbf{81.4}$_{\pm0.5}$ \\
CV-FM & ConvNeXt-XL & 64.4$_{\pm0.2}$ & 87.7$_{\pm0.0}$ & 71.7$_{\pm0.2}$ & 49.3$_{\pm0.2}$ & 59.0$_{\pm0.1}$ & 38.0$_{\pm0.1}$ & 67.8$_{\pm1.1}$ & 94.8$_{\pm0.3}$ & 41.6$_{\pm0.6}$ & 77.6$_{\pm0.8}$ \\
 & DINOv3 ConvNeXt-L & 65.8$_{\pm0.2}$ & 88.4$_{\pm0.0}$ & 71.8$_{\pm0.2}$ & 52.4$_{\pm0.1}$ & 60.1$_{\pm0.2}$ & 38.4$_{\pm0.1}$ & 69.8$_{\pm0.0}$ & \textbf{95.9}$_{\pm0.1}$ & 40.0$_{\pm0.9}$ & 77.1$_{\pm0.0}$ \\
 & DINOv3 ViT-L (LVD) & 66.6$_{\pm0.4}$ & \textbf{88.9}$_{\pm0.0}$ & \textbf{73.2}$_{\pm0.1}$ & 54.3$_{\pm0.3}$ & 61.3$_{\pm0.1}$ & 36.7$_{\pm0.0}$ & 71.5$_{\pm0.6}$ & 95.5$_{\pm0.3}$ & 42.7$_{\pm0.3}$ & 76.8$_{\pm0.8}$ \\
 & ResNet-50 & 58.7$_{\pm0.9}$ & 85.1$_{\pm0.4}$ & 69.7$_{\pm0.1}$ & 45.8$_{\pm0.3}$ & 55.9$_{\pm0.6}$ & 41.4$_{\pm0.1}$ & 59.3$_{\pm0.1}$ & 92.1$_{\pm0.2}$ & 37.4$_{\pm0.3}$ & 78.6$_{\pm0.3}$ \\
MidRes-FM & Prithvi-EO 2.0 600M & 60.7$_{\pm0.7}$ & 82.1$_{\pm0.1}$ & 65.1$_{\pm0.0}$ & 37.5$_{\pm0.5}$ & 52.1$_{\pm0.2}$ & 46.0$_{\pm0.2}$ & 41.9$_{\pm0.2}$ & 89.2$_{\pm0.8}$ & 21.9$_{\pm0.2}$ & 45.6$_{\pm0.7}$ \\
 & THOR v1 Base & 56.5$_{\pm0.2}$ & 78.9$_{\pm0.1}$ & 63.0$_{\pm0.0}$ & 32.3$_{\pm0.9}$ & 50.8$_{\pm0.3}$ & 46.6$_{\pm0.1}$ & 42.1$_{\pm0.0}$ & 88.1$_{\pm0.1}$ & 27.8$_{\pm0.5}$ & 49.1$_{\pm0.5}$ \\
 & TerraMind v1 L & 59.7$_{\pm0.4}$ & 84.3$_{\pm0.0}$ & 68.1$_{\pm0.1}$ & 43.7$_{\pm0.1}$ & 56.6$_{\pm0.2}$ & 40.4$_{\pm0.1}$ & 56.3$_{\pm0.0}$ & 87.9$_{\pm0.3}$ & 35.6$_{\pm0.1}$ & 71.0$_{\pm0.7}$ \\
\midrule
Baseline & ResNet-50 (ft)$^{*}$ & 63.8$_{\pm0.5}$ & 87.0$_{\pm0.9}$ & 71.7$_{\pm0.1}$ & 46.6$_{\pm0.5}$ & 61.0$_{\pm0.1}$ & 38.0$_{\pm0.2}$ & 58.1$_{\pm1.5}$ & 96.0$_{\pm0.1}$ & 44.2$_{\pm0.2}$ & 80.7$_{\pm0.1}$ \\
\bottomrule
\end{tabular}
}
\end{table*}

%% file: partials/table_loo.tex
\begin{table}[t]
\centering
\small
\setlength{\tabcolsep}{4pt}
\caption[Leave-one-out sensitivity]{Leave-one-out sensitivity after adding the continued model. $\rho$ is the Spearman correlation with the full 13-encoder ranking. The leader column names an encoder only where it differs from the full-suite leader at that fraction; a dash means unchanged. No encoder moves by more than one position.}
\label{tab:loo}
\begin{tabular}{@{}lrlrl@{}}
\toprule
& \multicolumn{2}{c}{10\%} & \multicolumn{2}{c}{100\%} \\
\cmidrule(lr){2-3}\cmidrule(lr){4-5}
Removed & $\rho$ & Leader & $\rho$ & Leader \\
\midrule
HOT & 0.995 & -- & 0.989 & -- \\
SN2 & 0.995 & -- & 0.995 & -- \\
SN3 & 0.995 & -- & 0.995 & -- \\
FLR & 0.995 & -- & 0.989 & -- \\
xV2 & 0.995 & -- & 0.995 & -- \\
OpC & 0.995 & -- & 1.000 & -- \\
PF & 0.989 & -- & 0.989 & -- \\
PV & 1.000 & -- & 0.989 & -- \\
EvW & 0.995 & -- & 0.995 & -- \\
NZC & 0.995 & LVD$\rightarrow$HR & 0.995 & -- \\
\bottomrule
\end{tabular}
\end{table}

%% file: sec/9_discussion.tex
\section{Discussion}\label{discussion}

\subsection{Interpretation of the findings}

The released-encoder benchmark does not reveal a general performance separation between natural-image and high-resolution remote-sensing pretraining. The first six positions are divided equally between CV-FM and HR-FM encoders, and each tier contains both leading and mid-ranked models. Strong web-image pretraining is therefore a competitive alternative to high-resolution remote-sensing pretraining for VHR RGB tasks. This conclusion concerns aggregate transfer performance, rather than equivalence between the two pretraining strategies: their per-dataset results show complementary strengths. 

SkySense provides the strongest result when labels are scarce, scoring 65.8 at the 10\% fraction compared with 64.0 for the best CV-FM, and it ranks first on the mean of the two fractions. At the full fraction, however, DINOv3 ViT-L (LVD) scores 69.5 and SkySense 69.2. The low-label result is consistent with the possibility that high-resolution remote-sensing pretraining improves sample efficiency, but the comparison cannot isolate the effect of pretraining data. SkySense uses a 655M-parameter hierarchical Swin-H backbone with 301.6 GFLOPs, whereas DINOv3 ViT-L has 303M parameters and 164.7 GFLOPs. SkySense also provides a native multi-scale feature hierarchy, while features from the single-scale ViT, like DINOv3, are resampled for the decoder. Architecture, capacity, compute, and pretraining data are therefore all potential contributors to its low-label advantage. 

The matched DINOv3 comparison gives more direct evidence about pretraining source. With architecture, training recipe and evaluation fixed, the web-pretrained LVD variant is slightly ahead in aggregate at both fractions. More specialized pretraining is thus not sufficient on its own to improve average VHR transfer. Nevertheless, the per-task results contain interesting information: at the full fraction, the SAT variant is stronger on canopy-height regression and both object-detection datasets, whereas LVD is stronger on segmentation, change-detection, and classification. This pattern suggests that satellite pretraining may preserve useful spatial or physically grounded cues while large-scale web pretraining supports broader semantic discrimination. 

Several properties of the pretraining corpora could explain the aggregate advantage of LVD, but the present experiments do not distinguish among them. LVD-1689M contains approximately 3.4 times as many images as SAT-493M and is likely to cover substantially greater visual and semantic diversity. SAT-493M, on the other hand, consists of imagery from one provider at a single nominal resolution. The observed difference could therefore reflect corpus size, visual diversity, sensor diversity, or their interaction, rather than a disadvantage of satellite imagery itself. Isolating the different factors would require controlled pretraining experiments with matched data volumes and training budgets. 

The MidRes-FM encoders occupy the final three positions, providing the clearest tier-level pattern in the benchmark. This supports the hypothesized difficulty of transferring representations learned on multi-spectral inputs and at Sentinel or Landsat scale to RGB imagery in which small individual objects and fine spatial boundaries are resolved. These encoders are designed for multispectral inputs but receive only RGB with their remaining channels set to zero, and their backbones are frozen. The experiments therefore show that the evaluated Sentinel-scale encoders transfer least effectively when repurposed as frozen RGB encoders for VHR imagery, not that medium-resolution pretraining is intrinsically ineffective in its native modality. General benchmarks like PANGAEA~\cite{pangaea}, GEO-Bench~\cite{geobench1} and GEO-Benchand 2~\cite{geobench2} are instead the right tool to evaluate MOdRes-FM encoders in their native modalities.


\subsection{Generalize, then specialize}

The continued-pretraining experiment turns the released-model comparison into a design test. Starting from the strongest web encoder and continuing the same self-supervised objective on VHR imagery improves its aggregate at both fractions, with the clearest gain when labels are scarce. The continued model obtains the highest observed full-fraction score and remains second to SkySense at 10\% labels, without changing its inference cost. Broad visual pretraining is therefore not merely a competing alternative to VHR pretraining; it is also a useful initialization for it.

The ordering of the three DINOv3 checkpoints suggests that the sequence of data
regimes matters. Training on the specialized SAT-493M corpus alone does not exceed
the web model in aggregate, whereas beginning from LVD-1689M and then specializing
on a smaller, multi-source VHR corpus does. The most plausible interpretation is
that the first stage learns broadly transferable semantic and visual structure and
the second adjusts it to VHR texture, scale, and acquisition statistics. DINOv3's
original web pretraining was already extended at scale until downstream performance
saturated~\cite{dinov3}; the observed movement after switching to VHR data is
therefore more naturally attributed to the new domain than to another short block
of unchanged web training. The CKA and ImageNet-retention results support this
reading: early features and general recognition performance remain stable while
the deepest layers move.

The combination is nevertheless incomplete. At the full fraction, DINOv3-SAT
retains the lowest canopy-height error and remains ahead of the continued model on
both detection datasets; SkySense also remains the strongest detector. Continued
pretraining improves the aggregate mainly by preserving already strong segmentation
features, improving land-cover and tree-species recognition, and closing part of
the low-label detection gap. It does not introduce every spatial or physically
grounded prior learned by VHR-native models. A future continuation corpus or
objective designed explicitly for geometry, scale, or sensor consistency may be
needed to capture those remaining strengths.

\subsection{Benchmark and model-development considerations}

The Open-Canopy-Small result requires caution. DINOv3-SAT is the only released encoder
to lead this dataset at both fractions, and Open-Canopy was also used as a
downstream benchmark in the original DINOv3 study~\cite{dinov3}. This creates
model-development overlap but does not establish leakage: the DINOv3 paper reports
a fixed satellite-pretraining schedule and does not state that Open-Canopy selected
the released checkpoint. The result should therefore be read as potentially
affected by prior benchmark exposure rather than explained by it.

The continued model introduces a related but transparent form of benchmark-guided
development. VHR-Bench motivates the choice of initialization and a four-task
development subset informs the training-length analysis. The continued-pretraining
objective itself uses no labels, and known image intersections between its corpus
and VHR-Bench are excluded. We report the full per-dataset tables (\cref{tab:results100,tab:results10}) so that gains on
development tasks and the remaining tasks can be distinguished. Stronger tile-level
provenance across all broad imagery archives would further reduce the possibility
of residual geographic overlap.

\subsection{Limitations}

The pretraining tiers are descriptive groupings rather than controlled experimental conditions. Models within and across tiers differ in architecture, training objective, capacity, pretraining volume, modalities and sensor coverage. Except for the two DINOv3 ViT-L variants, performance differences cannot be attributed solely to pretraining resolution or domain. Controlled comparisons that vary one property at a time would be required to separate the effects of resolution, data scale, architecture, and modality. 

The study is restricted to RGB optical imagery. Near-infrared, multispectral, and VHR SAR inputs are outside its scope. This restriction particularly affects the MidRes-FM encoders, which receive zero-filled non-RGB channels, and may also limit the forestry tasks, for which near-infrared information is informative. The conclusions therefore characterize transfer to RGB VHR imagery rather than the full capabilities of the evaluated encoders.

The benchmark spans five task families but is not numerically balanced among them: four datasets address semantic segmentation and a fifth addresses a very similar task, pixel-level change detection. Classification, detection and regression consequently have narrower task and geographic coverage. The detection datasets both concern small animals over relatively uniform backgrounds and should not be treated as representative of vehicle, ship, or building detection. Likewise, classification is limited to France and Germany, regression to France, and detection to the United States and New Zealand. These limitations primarily reflect the scarcity of openly licensed VHR datasets.

The evaluation uses a frozen backbone, one predetermined decoder per task family, and a fixed training budget of 80 epochs. This supports a uniform comparison of representation quality but does not measure the models' performance after end-to-end finetuning, their convergence speed, or their maximum possible performance with an architecture-tailored decoder design. Native hierarchical features may also interact differently with the selected decoders than resampled single-scale transformer features.

Finally, the aggregate assigns equal value to every dataset, irrespective of dataset size. This prevents large datasets from dominating the benchmark but allows highly discriminative datasets to affect the ordering of closely matched encoders. The leave-one-out analysis shows that the ranking is generally stable, with only the full-fraction leader changing when PureForest is removed. Conversely, the near-saturation of m-pv4ger gives that dataset little ability to distinguish the strongest encoders. Future benchmark versions should retain per-dataset results alongside the aggregate and consider replacing or extending tasks that no longer separate state-of-the-art models.

%% file: sec/11_conclusion.tex
\section{Conclusion}\label{conclusion}

We introduced VHR-Bench, a benchmark for evaluating visual encoders on RGB very-high-resolution Earth observation imagery. The suite spans ten datasets, five task families, eight application domains, and both satellite and aerial platforms. Using a common frozen-backbone protocol, we evaluated twelve encoders at two label fractions to examine how pretraining domain and resolution relate to transfer at VHR.

Our results show that high-resolution remote-sensing and general-purpose computer-vision encoders both provide strong representations for VHR tasks but exhibit complementary task profiles, whereas encoders pretrained only at a Sentinel or Landsat scale transfer least effectively under RGB input. The matched DINOv3 comparison makes the complementarity especially visible: web pretraining is stronger in aggregate and on semantic tasks, while satellite pretraining retains advantages on canopy-height regression and object detection.

Guided by this diagnostic, we continued the self-supervised training of the web-pretrained DINOv3 ViT-L on 1.02 million VHR tiles and returned the resulting encoder to the benchmark. The continued model improved on its initialization and obtains the highest observed full-fraction aggregate, and ranks second only to SkySense at 10\% labels. Because only the weights change, the gain adds no inference cost. Representation analysis further shows that the early layers and ImageNet recognition are preserved while the deepest layers specialize.  

Overall, VHR-specific pretraining is not required to obtain strong performance on the tasks considered here, but neither is general web-image pretraining uniformly preferable. The strongest practical direction may therefore be to combine the scale and visual diversity of natural-image pretraining with the spatial, sensor, and physical priors available in high-resolution Earth observation imagery. As a first step in this direction, a short label-free run of continual VHR pretraining specializes a web-pretrained encoder, with the largest benefit when labels are scarce. VHR-Bench provides a reproducible basis for evaluating such models and for measuring whether future pretraining strategies improve performance consistently across VHR applications rather than on a narrow set of tasks.

%% file: references.bib
@IEEEtranBSTCTL{BSTctl,
  CTLdash_repeated_names = "no"
}

@article{geobench2,
  author = {Simumba, Naomi and Lehmann, Nils and Fraccaro, Paolo and Alemohammad, Hamed and De Mel, Geeth and Khan, Salman and Maskey, Manil and Longepe, Nicolas and Zhu, Xiao Xiang and Kerner, Hannah and Bernabe-Moreno, Juan and Lacoste, Alexandre},
  title = {{GEO-Bench-2: From Performance to Capability, Rethinking Evaluation in Geospatial AI}},
  journal = {arXiv preprint arXiv:2511.15658},
  year = {2025}
}

@inproceedings{fmow,
  author = {Christie, Gordon and Fendley, Neil and Wilson, James and Mukherjee, Ryan},
  title = {{Functional Map of the World}},
  booktitle = {2018 IEEE/CVF Conference on Computer Vision and Pattern Recognition (CVPR)},
  year = {2018},
  doi = {10.1109/CVPR.2018.00646}
}

@inproceedings{worldstrat,
  author = {Cornebise, Julien and Or{\v{s}}oli{\'c}, Ivan and Kalaitzis, Freddie},
  title = {{Open High-Resolution Satellite Imagery: The WorldStrat Dataset -- With Application to Super-Resolution}},
  booktitle = {Advances in Neural Information Processing Systems 35 (NeurIPS 2022) Datasets and Benchmarks Track},
  year = {2022}
}

@article{dofa,
  author = {Xiong, Zhitong and Wang, Yi and Zhang, Fahong and Stewart, Adam J. and Hanna, Joëlle and Borth, Damian and Papoutsis, Ioannis and Saux, Bertrand Le and Camps-Valls, Gustau and Zhu, Xiao Xiang},
  title = {{Neural Plasticity-Inspired Multimodal Foundation Model for Earth Observation}},
  journal = {arXiv preprint arXiv:2403.15356},
  year = {2024}
}

@article{dinov3,
  author = {Siméoni, Oriane and Vo, Huy V. and Seitzer, Maximilian and Baldassarre, Federico and Oquab, Maxime and Jose, Cijo and Khalidov, Vasil and Szafraniec, Marc and Yi, Seungeun and Ramamonjisoa, Michaël and Massa, Francisco and Haziza, Daniel and Wehrstedt, Luca and Wang, Jianyuan and Darcet, Timothée and Moutakanni, Théo and Sentana, Leonel and Roberts, Claire and Vedaldi, Andrea and Tolan, Jamie and Brandt, John and Couprie, Camille and Mairal, Julien and Jégou, Hervé and Labatut, Patrick and Bojanowski, Piotr},
  title = {{DINOv3}},
  journal = {arXiv preprint arXiv:2508.10104},
  year = {2025}
}

@article{strahler1986,
  author = {Strahler, Alan H. and Woodcock, Curtis E. and Smith, James A.},
  title = {{On the nature of models in remote sensing}},
  journal = {Remote Sensing of Environment},
  volume = {20},
  number = {2},
  pages = {121–139},
  year = {1986},
  month = {Oct},
  publisher = {Elsevier BV},
  doi = {10.1016/0034-4257(86)90018-0},
  issn = {0034-4257}
}

@article{thor,
  author = {Forgaard, Theodor and Reksten, Jarle H. and Waldeland, Anders U. and Marsocci, Valerio and Longépé, Nicolas and Kampffmeyer, Michael and Salberg, Arnt-Børre},
  title = {{THOR: A Versatile Foundation Model for Earth Observation Climate and Society Applications}},
  journal = {arXiv preprint arXiv:2601.16011},
  year = {2026}
}

@misc{clay,
  author = {{Clay Foundation}},
  title = {{Clay Foundation Model v1.5}},
  year = {2024},
  howpublished = {\url{https://clay-foundation.github.io/model/release-notes/specification.html}},
  note = {Open masked-autoencoder Vision Transformer geospatial foundation model}
}

@inproceedings{skysense,
  author = {Guo, Xin and Lao, Jiangwei and Dang, Bo and Zhang, Yingying and Yu, Lei and Ru, Lixiang and Zhong, Liheng and Huang, Ziyuan and Wu, Kang and Hu, Dingxiang and He, Huimei and Wang, Jian and Chen, Jingdong and Yang, Ming and Zhang, Yongjun and Li, Yansheng},
  title = {{SkySense: A Multi-Modal Remote Sensing Foundation Model Towards Universal Interpretation for Earth Observation Imagery}},
  booktitle = {2024 IEEE/CVF Conference on Computer Vision and Pattern Recognition (CVPR)},
  pages = {27662–27673},
  year = {2024},
  month = {June},
  publisher = {IEEE},
  doi = {10.1109/cvpr52733.2024.02613}
}

@inproceedings{terramind,
  author = {Jakubik, Johannes and Yang, Felix and Blumenstiel, Benedikt and Scheurer, Erik and Sedona, Rocco and Maurogiovanni, Stefano and Bosmans, Jente and Dionelis, Nikolaos and Marsocci, Valerio and Kopp, Niklas and Ramachandran, Rahul and Fraccaro, Paolo and Brunschwiler, Thomas and Cavallaro, Gabriele and Bernabe-Moreno, Juan and Longépé, Nicolas},
  title = {{TerraMind: Large-Scale Generative Multimodality for Earth Observation}},
  booktitle = {2025 IEEE/CVF International Conference on Computer Vision (ICCV)},
  pages = {7383–7394},
  year = {2025},
  month = {Oct},
  publisher = {IEEE},
  doi = {10.1109/iccv51701.2025.00693}
}

@inproceedings{geobench1,
  author = {Lacoste, Alexandre and Lehmann, Nils and Rodriguez, Pau and Sherwin, Evan and Kerner, Hannah and Lütjens, Björn and Irvin, Jeremy and Dao, David and Alemohammad, Hamed and Drouin, Alexandre and Gunturkun, Mehmet and Huang, Gabriel and Vazquez, David and Newman, Dava and Bengio, Yoshua and Ermon, Stefano and Zhu, Xiaoxiang},
  title = {{GEO-Bench: Toward Foundation Models for Earth Monitoring}},
  booktitle = {Advances in Neural Information Processing Systems (NeurIPS) 36},
  pages = {51080–51093},
  year = {2023},
  doi = {10.52202/075280-2223},
}

@article{prithvi2,
  author = {Szwarcman, Daniela and Roy, Sujit and Fraccaro, Paolo and Gíslason, {\TH}orsteinn Elí and Blumenstiel, Benedikt and Ghosal, Rinki and de Oliveira, Pedro Henrique and de Sousa Almeida, Joao Lucas and Sedona, Rocco and Kang, Yanghui and Chakraborty, Srija and Wang, Sizhe and Gomes, Carlos and Kumar, Ankur and Gaur, Vishal and Truong, Myscon and Godwin, Denys and Khallaghi, Sam and Lee, Hyunho and Hsu, Chia-Yu and Asanjan, Ata Akbari and Mujeci, Besart and Shidham, Disha and Balogun, Rufai Omowunmi and Kolluru, Venkatesh and Keenan, Trevor and Arevalo, Paulo and Li, Wenwen and Alemohammad, Hamed and Olofsson, Pontus and Mayer, Timothy and Hain, Christopher and Kennedy, Robert and Zadrozny, Bianca and Bell, David and Cavallaro, Gabriele and Watson, Campbell and Maskey, Manil and Ramachandran, Rahul and Moreno, Juan Bernabe},
  title = {{Prithvi-EO-2.0: A Versatile Multitemporal Foundation Model for Earth Observation Applications}},
  journal = {IEEE Transactions on Geoscience and Remote Sensing},
  volume = {64},
  pages = {1–20},
  year = {2026},
  publisher = {Institute of Electrical and Electronics Engineers (IEEE)},
  doi = {10.1109/tgrs.2025.3642610},
  issn = {1558-0644}
}

@article{pangaea,
  author = {Marsocci, Valerio and Jia, Yuru and Bellier, Georges Le and Kerekes, David and Zeng, Liang and Hafner, Sebastian and Gerard, Sebastian and Brune, Eric and Yadav, Ritu and Shibli, Ali and Fang, Heng and Ban, Yifang and Vergauwen, Maarten and Audebert, Nicolas and Nascetti, Andrea},
  title = {{PANGAEA: Assessing Geospatial Foundation Models Capabilities through a Global and Inclusive Benchmark}},
  journal = {IEEE Geoscience and Remote Sensing Magazine},
  volume = {14},
  number = {1},
  pages = {245–285},
  year = {2026},
  month = {Feb},
  publisher = {Institute of Electrical and Electronics Engineers (IEEE)},
  doi = {10.1109/mgrs.2025.3628194},
  issn = {2473-2397}
}

@article{bommasani2021,
  author = {Bommasani, Rishi and Hudson, Drew A. and Adeli, Ehsan and Altman, Russ and Arora, Simran and von Arx, Sydney and Bernstein, Michael S. and Bohg, Jeannette and Bosselut, Antoine and Brunskill, Emma and Brynjolfsson, Erik and Buch, Shyamal and Card, Dallas and Castellon, Rodrigo and Chatterji, Niladri and Chen, Annie and Creel, Kathleen and Davis, Jared Quincy and Demszky, Dora and Donahue, Chris and Doumbouya, Moussa and Durmus, Esin and Ermon, Stefano and Etchemendy, John and Ethayarajh, Kawin and Fei-Fei, Li and Finn, Chelsea and Gale, Trevor and Gillespie, Lauren and Goel, Karan and Goodman, Noah and Grossman, Shelby and Guha, Neel and Hashimoto, Tatsunori and Henderson, Peter and Hewitt, John and Ho, Daniel E. and Hong, Jenny and Hsu, Kyle and Huang, Jing and Icard, Thomas and Jain, Saahil and Jurafsky, Dan and Kalluri, Pratyusha and Karamcheti, Siddharth and Keeling, Geoff and Khani, Fereshte and Khattab, Omar and Koh, Pang Wei and Krass, Mark and Krishna, Ranjay and Kuditipudi, Rohith and Kumar, Ananya and Ladhak, Faisal and Lee, Mina and Lee, Tony and Leskovec, Jure and Levent, Isabelle and Li, Xiang Lisa and Li, Xuechen and Ma, Tengyu and Malik, Ali and Manning, Christopher D. and Mirchandani, Suvir and Mitchell, Eric and Munyikwa, Zanele and Nair, Suraj and Narayan, Avanika and Narayanan, Deepak and Newman, Ben and Nie, Allen and Niebles, Juan Carlos and Nilforoshan, Hamed and Nyarko, Julian and Ogut, Giray and Orr, Laurel and Papadimitriou, Isabel and Park, Joon Sung and Piech, Chris and Portelance, Eva and Potts, Christopher and Raghunathan, Aditi and Reich, Rob and Ren, Hongyu and Rong, Frieda and Roohani, Yusuf and Ruiz, Camilo and Ryan, Jack and Ré, Christopher and Sadigh, Dorsa and Sagawa, Shiori and Santhanam, Keshav and Shih, Andy and Srinivasan, Krishnan and Tamkin, Alex and Taori, Rohan and Thomas, Armin W. and Tramèr, Florian and Wang, Rose E. and Wang, William and Wu, Bohan and Wu, Jiajun and Wu, Yuhuai and Xie, Sang Michael and Yasunaga, Michihiro and You, Jiaxuan and Zaharia, Matei and Zhang, Michael and Zhang, Tianyi and Zhang, Xikun and Zhang, Yuhui and Zheng, Lucia and Zhou, Kaitlyn and Liang, Percy},
  title = {{On the Opportunities and Risks of Foundation Models}},
  journal = {arXiv preprint arXiv:2108.07258},
  year = {2021}
}

@article{Xiao_2025,
  author = {Xiao, Aoran and Xuan, Weihao and Wang, Junjue and Huang, Jiaxing and Tao, Dacheng and Lu, Shijian and Yokoya, Naoto},
  title = {{Foundation Models for Remote Sensing and Earth Observation: A survey}},
  journal = {IEEE Geoscience and Remote Sensing Magazine},
  volume = {13},
  number = {4},
  pages = {297–324},
  year = {2025},
  month = {Dec},
  publisher = {Institute of Electrical and Electronics Engineers (IEEE)},
  doi = {10.1109/mgrs.2025.3576766},
  issn = {2473-2397}
}

@inproceedings{eosfm,
  author = {Adorni, Pierre and Pham, Minh-Tan and May, Stéphane and Lefèvre, Sébastien},
  title = {{EoS-FM: Can an Ensemble of Specialist Models act as a Generalist Feature Extractor?}},
  booktitle={European Conference on Computer Vision (ECCV)},
  year = {2026}
}

@article{van2018spacenet,
  title={Spacenet: A remote sensing dataset and challenge series},
  author={Van Etten, Adam and Lindenbaum, Dave and Bacastow, Todd M},
  journal={arXiv preprint arXiv:1807.01232},
  year={2018}
}

@inproceedings{gupta2019creating,
  title={Creating xBD: A dataset for assessing building damage from satellite imagery},
  author={Gupta, Ritwik and Goodman, Bryce and Patel, Nirav and Hosfelt, Richard and Sajeev, Sandra and Heim, Eric and Doshi, Jigar and Lucas, Keane and Choset, Howard and Gaston, Matthew},
  year={2019},
  booktitle={Proceedings of the IEEE/CVF Conference on Computer Vision and Pattern Recognition (CVPR) Workshops},
  publisher={Carnegie Mellon University}
}

@inproceedings{ign2023flair2,
      title={FLAIR: a Country-Scale Land Cover Semantic Segmentation Dataset From Multi-Source Optical Imagery}, 
      author={Anatol Garioud and Nicolas Gonthier and Loic Landrieu and Apolline De Wit and Marion Valette and Marc Poupée and Sébastien Giordano and Boris Wattrelos},
      year={2023},
      booktitle={Advances in Neural Information Processing Systems (NeurIPS) 2023},
      doi={https://doi.org/10.48550/arXiv.2310.13336},
}

@article{garner2024everwatch,
  title={EverWatch benchmark: training and evalutation data for detection and species classification of Everglades wading birds from airborne imagery},
  author={Garner, L and Weinstein, B and Rickershauser, M and Baldino, M and Coates, H and Commins, M and Faber, T and Gula, J and Van Ert, S and White, E and others},
  journal={Zenodo},
  year={2024}
}

@dataset{nz_cattle,
  author       = {Dr. Diab Abuaiadah and
                  Alexander Switzer},
  title        = {New Zealand Cattle Detection},
  month        = jan,
  year         = 2022,
  publisher    = {Zenodo},
  version      = {1.0.0},
  doi          = {10.5281/zenodo.5908869},
  url          = {https://doi.org/10.5281/zenodo.5908869},
}

@article{mayer20223d,
  title={3D-PV-Locator: Large-scale detection of rooftop-mounted photovoltaic systems in 3D},
  author={Mayer, Kevin and Rausch, Benjamin and Arlt, Marie-Louise and Gust, Gunther and Wang, Zhecheng and Neumann, Dirk and Rajagopal, Ram},
  journal={Applied Energy},
  volume={310},
  pages={118469},
  year={2022},
  publisher={Elsevier}
}

@inproceedings{gaydon2025pureforest,
  title={Pureforest: A large-scale aerial lidar and aerial imagery dataset for tree species classification in monospecific forests},
  author={Gaydon, Charles and Roche, Floryne},
  booktitle={2025 IEEE/CVF Winter Conference on Applications of Computer Vision (WACV)},
  pages={5895--5904},
  year={2025},
  organization={IEEE}
}

@misc{hot_building_segmentation_2026,
  title={HOT Building Segmentation Dataset},
  author={Humanitarian OpenStreetMap Team},
  year={2026},
  publisher={Hugging Face},
  howpublished={\url{https://huggingface.co/datasets/hotosm/vhr-building-segmentation}}
}

@inproceedings{resnet,
  title={Deep residual learning for image recognition},
  author={He, Kaiming and Zhang, Xiangyu and Ren, Shaoqing and Sun, Jian},
  booktitle={Proceedings of the IEEE Conference on Computer Vision and Pattern Recognition (CVPR)},
  pages={770--778},
  year={2016}
}

@inproceedings{convnext,
  author = {Liu, Zhuang and Mao, Hanzi and Wu, Chao-Yuan and Feichtenhofer, Christoph and Darrell, Trevor and Xie, Saining},
  title = {{A ConvNet for the 2020s}},
  booktitle = {2022 IEEE/CVF Conference on Computer Vision and Pattern Recognition (CVPR)},
  pages = {11966–11976},
  year = {2022},
  month = {June},
  publisher = {IEEE},
  doi = {10.1109/cvpr52688.2022.01167}
}

@inproceedings{fogel2024opencanopy,
  author = {Fogel, Fajwel and Perron, Yohann and Besic, Nikola and Saint-André, Laurent and Pellissier-Tanon, Agnès and Schwartz, Martin and Boudras, Thomas and Fayad, Ibrahim and d’Aspremont, Alexandre and Landrieu, Loic and Ciais, Philippe},
  title = {{Open-Canopy: Towards Very High Resolution Forest Monitoring}},
  booktitle = {2025 IEEE/CVF Conference on Computer Vision and Pattern Recognition (CVPR)},
  pages = {1395–1406},
  year = {2025},
  month = {June},
  publisher = {IEEE},
  doi = {10.1109/cvpr52734.2025.00138}
}

@article{fino,
  title={Who Needs Labels? Adapting Vision Foundation Models With the Metadata You Already Have},
  author={Gard{\`e}s, Elouan and Yi, Seung Eun and Ahuja, Kartik and Moutakanni, Th{\'e}o and Vo, Huy V and Bojanowski, Piotr and Pernice, Wolfgang M and Landrieu, Lo{\"\i}c and Couprie, Camille},
  journal={arXiv preprint arXiv:2606.05107},
  year={2026}
}

@inproceedings{
explora,
title={Ex{PL}o{RA}: Parameter-Efficient Extended Pre-Training to Adapt Vision Transformers under Domain Shifts},
author={Samar Khanna and Medhanie Irgau and David B. Lobell and Stefano Ermon},
booktitle={Forty-second International Conference on Machine Learning},
year={2025},
url={https://openreview.net/forum?id=OtxLhobhwb}
}

@inproceedings{kornblith2019similarity,
  title={Similarity of neural network representations revisited},
  author={Kornblith, Simon and Norouzi, Mohammad and Lee, Honglak and Hinton, Geoffrey},
  booktitle={International conference on machine learning},
  pages={3519--3529},
  year={2019},
  organization={PMLR}
}

@inproceedings{satlas,
  title={Satlaspretrain: A large-scale dataset for remote sensing image understanding},
  author={Bastani, Favyen and Wolters, Piper and Gupta, Ritwik and Ferdinando, Joe and Kembhavi, Aniruddha},
  booktitle={2023 IEEE/CVF international conference on computer vision (ICCV)},
  pages={16726--16736},
  year={2023},
  organization={IEEE}
}

@misc{maxar_open_data,
  author       = {{Maxar}},
  title        = {Maxar Open Data Program},
  howpublished = {\url{https://registry.opendata.aws/maxar-open-data/}},
  note         = {Accessed: Aug. 24, 2026}
}

@misc{openaerialmap,
  author       = {{Humanitarian OpenStreetMap Team}},
  title        = {OpenAerialMap},
  howpublished = {\url{https://openaerialmap.org/}},
  note         = {Accessed: Aug. 24, 2026}
}

@misc{naip,
  author    = {{Earth Resources Observation and Science (EROS) Center}},
  title     = {National Agriculture Imagery Program: 2003--present},
  year      = {2023},
  publisher = {U.S. Geological Survey},
  doi       = {10.5066/F7QN651G},
  howpublished = {\url{https://naip-usdaonline.hub.arcgis.com/}},
  note = {Accessed: Sept. 2, 2026}
}

@misc{ramp,
  author       = {{DevGlobal Partners}},
  title        = {Replicable {AI} for Microplanning ({RAMP}): Data Labeling Specifications},
  year         = {2022},
  howpublished = {\url{https://rampml.global/}},
  note         = {Accessed: Aug. 24, 2026}
}

@misc{planet_disaster_data,
  author       = {{Planet Labs PBC}},
  title        = {Planet Disaster Data},
  year         = {2026},
  howpublished = {\url{https://www.planet.com/disasterdata/}},
  note         = {Accessed: Aug. 24, 2026}
}

@inproceedings{tessera,
  title={Tessera: Temporal embeddings of surface spectra for earth representation and analysis},
  author={Feng, Zhengpeng and Atzberger, Clement and Jaffer, Sadiq and Knezevic, Jovana and Sormunen, Silja and Young, Robin and Lisaius, Madeline C and Immitzer, Markus and Jackson, Toby and Ball, James and others},
  booktitle={Proceedings of the IEEE/CVF Conference on Computer Vision and Pattern Recognition},
  pages={34818--34831},
  year={2026}
}

@article{alphaearth,
  title={Alphaearth foundations: An embedding field model for accurate and efficient global mapping from sparse label data},
  author={Brown, Christopher F and Kazmierski, Michal R and Pasquarella, Valerie J and Rucklidge, William J and Samsikova, Masha and Zhang, Chenhui and Shelhamer, Evan and Lahera, Estefania and Wiles, Olivia and Ilyushchenko, Simon and others},
  journal={arXiv preprint arXiv:2507.22291},
  year={2025}
}

@inproceedings{panopticon,
    author={Waldmann, Leonard and Shah, Ando and Wang, Yi and Lehmann, Nils and Stewart, Adam and Xiong, Zhitong and Zhu, Xiao Xiang and Bauer, Stefan and Chuang, John},
    title={Panopticon: Advancing Any-Sensor Foundation Models for Earth Observation},
    booktitle={Proceedings of the Computer Vision and Pattern Recognition Conference (CVPR) Workshops},
    year={2025},
    pages={2204-2214}
}
